\documentclass[sigconf]{acmart}
\usepackage{booktabs} % For formal tables
\usepackage{epigraph}
\usepackage{algorithm}
\usepackage{algpseudocode}
\usepackage{graphicx}
\usepackage{multirow}
\usepackage{amsfonts,amssymb,amsmath,color}
\usepackage{url}
\usepackage{subfigure}
\usepackage{todonotes}
\usepackage[shortlabels]{enumitem}

\def\0{{\bf 0}}
\def\1{{\bf 1}}

\def\EM{{\mathcal E}}

\def\HM{{\mathcal H}}

\def\LM{{\mathcal L}}

\def\PM{{\mathcal P}}

\def\RM{{\mathcal R}}

\def\TM{{\mathcal T}}

\def\VM{{\mathcal V}}

\newcommand{\nop}[1]{}

\newtheorem{definition}{Definition}

\copyrightyear{2020}
\acmYear{2020}
\setcopyright{iw3c2w3}
\acmConference[WWW '20]{Proceedings of The Web Conference 2020}{April 20--24, 2020}{Taipei, Taiwan}
\acmBooktitle{Proceedings of The Web Conference 2020 (WWW '20), April 20--24, 2020, Taipei, Taiwan}
\acmPrice{}
\acmDOI{10.1145/3366423.3380107}
\acmISBN{978-1-4503-7023-3/20/04}

\begin{document}
\title{ASER: A Large-scale Eventuality Knowledge Graph}
% \author{Anonymous Authors}
\author{Hongming Zhang}
\authornote{Authors contributed equally to this research.}
\email{hzhangal@cse.ust.hk}
\affiliation{%
  \institution{CSE, HKUST}
  \city{Hong Kong}
}
\author{Xin Liu}
\authornotemark[1]
\email{xliucr@cse.ust.hk}
\affiliation{%
  \institution{CSE, HKUST}
  \city{Hong Kong}
}
\author{Haojie Pan}
\authornotemark[1]
\email{hpanad@cse.ust.hk}
\affiliation{%
  \institution{CSE, HKUST}
  \city{Hong Kong}
}
\author{Yangqiu Song}
\email{yqsong@cse.ust.hk}
\affiliation{%
  \institution{CSE, HKUST}
  \city{Hong Kong}
}
\author{Cane Wing-Ki  Leung}
\email{caneleung@wisers.com}
\affiliation{%
  \institution{Wisers AI Lab}
  \city{Hong Kong}
}
% \author{\name Hongming Zhang\thanks{Equal contribution.} \email hzhangal@cse.ust.hk \\
%       \name Xin Liu$^{*}$ \email xliucr@cse.ust.hk \\
%       \name Haojie Pan$^{*}$ \email hpanad@cse.ust.hk \\
%       \name Yangqiu Song \email yqsong@cse.ust.hk \\
%       \addr CSE, HKUST, Hong Kong
%       \AND
%       \name Cane Wing-Ki  Leung \email caneleung@wisers.com \\
%       \addr Wisers AI Lab, Hong Kong} 
       
\begin{abstract}
Understanding human's language requires complex world knowledge.
However, existing large-scale knowledge graphs mainly focus on knowledge about entities while ignoring knowledge about activities, states, or events, which are used to describe how entities or things act in the real world.
To fill this gap, we develop ASER (activities, states, events, and their relations), a large-scale eventuality knowledge graph extracted from more than 11-billion-token unstructured textual data.
ASER contains 15 relation types belonging to five categories, 194-million unique eventualities, and 64-million unique edges among them.
Both intrinsic and extrinsic evaluations demonstrate the quality and effectiveness of ASER.

\end{abstract}  

\maketitle

\section{Introduction}\label{sec:introduction}

% \epigraph{\textit{`The most incomprehensible thing about the world is that it is at all comprehensible.'} }{\textbf{Albert Einstein}}

In his conceptual semantics theory, Ray Jackendoff, a Rumelhart Prize\footnote{The David E. Rumelhart Prize is funded for contributions to the theoretical foundations of human cognition.} winner, describes semantic meaning as `a finite set of mental primitives and a finite set of principles of mental combination~\cite{Jackendoff}'. The primitive units of semantic meanings include {\it Thing} (or {\it Object}), {\it Activity}\footnote{In his original book, he called it {\it Action}. But given the other definitions and terminologies we adopted~\cite{ALEXANDER1978,bach1986algebra}, it means {\it Activity}.}, {\it State}, {\it Event}, {\it Place}, {\it Path}, {\it Property}, {\it Amount}, etc.
Understanding the semantics related to the world requires the understanding of these units and their relations.
Traditionally, linguists and domain experts built knowledge graphs (KGs)\footnote{Traditionally, people used the term `knowledge base' to describe the database containing human knowledge. In 2012, Google released its knowledge graph where vertices and edges in a knowledge base are emphasized. We discuss in the context of the knowledge graph, as our knowledge is also constructed as a complex graph. For more information about terminologies, please refer to~\cite{DBLP:conf/i-semantics/EhrlingerW16}.} to formalize these units and enumerate categories (or senses) and relations of them.
Typical KGs include WordNet~\cite{WordNet} for words, FrameNet~\cite{framenet} for events, and Cyc~\cite{researchCyc} and ConceptNet~~\cite{liu2004conceptnet} for commonsense knowledge. 
However, their small scales restricted their usage in real-world applications.

Nowadays, with the growth of Web contents, computational power, and the availability of crowdsourcing platforms, many modern and large-scale KGs, such as Freebase~\cite{freebase}, KnowItAll~\cite{knowitall}, TextRunner~\cite{BankoCSBE07}, YAGO~\cite{YAGO}, DBpedia~\cite{auer2007dbpedia}, NELL~\cite{NELL}, Probase~\cite{wu2011taxonomy}, and Google Knowledge Vault~\cite{dong2014knowledge}, have been built based on semi-automatic mechanisms.
Most of these KGs are designed and constructed based on {\it Things} or {\it Objects}, such as instances and their concepts, named entities and their categories, as well as their properties and relations.
On top of them, a lot of semantic understanding problems such as question answering~\cite{BerantCFL13} can be supported by grounding natural language texts on knowledge graphs, e.g., asking a bot for {the nearest restaurants for lunch}.
Nevertheless, these KGs may fall short in circumstances that require not only knowledge about {\it Things} or {\it Objects}, but also those about {\it Activities}, {\it States}, and {\it Events}.
%However, important units including {\it State}, {\it Activity}, and {\it Event} are missing, which can affect many semantic understanding problems.
Consider the following utterance that a human would talk to the bot: {`I am hungry'}, which may also imply one's need for restaurant recommendation. This, however, will not be possible unless the bot is able to identify that the consequence of being hungry would be {`having lunch'} at noon.
%Here {`being hungry'} is a {\it state} and {`having lunch'} is an {\it action}.
%If a bot can build a relation between them in a knowledge graph,  can make possible suggestions when identifying the need of a human
%It turns out that understanding a user's intent requires the knowledge about the relation between `being hungry' and `having lunch'.

In this paper, we propose an approach to discovering useful real-world knowledge about {\it Activities} (or process, e.g., `I sleep'), {\it States} (e.g., 'I am hungry'), {\it Events} (e.g., `I make a call'), and their {\it Relations} (e.g., `I am hungry' may result in `I have lunch'), for which we call ASER. 
In fact,  {\it Activities}, {\it States}, and {\it Events}, which are expressed by verb-related clauses, are all eventualities following the commonly adopted terminology and categorization proposed by Mourelatos~\cite{ALEXANDER1978} and Bach~\cite{bach1986algebra}.
While both activity and event are occurrences (actions) described by active verbs, a state is usually described by a stative verb and cannot be qualified as actions. The difference between an activity and an event is that an event is defined as an occurrence that is inherently countable~\cite{ALEXANDER1978}. For example, `The coffee machine brews a cup of coffee once more' is an event because it admits a countable noun `a cup' and cardinal count adverbials `once', while `The coffee machine brews coffee' is not an event with an imperfective aspect and it is not countable.
Thus, ASER is essentially an eventuality-centric knowledge graph. 

\begin{figure}
\centering
\includegraphics[width=\linewidth]{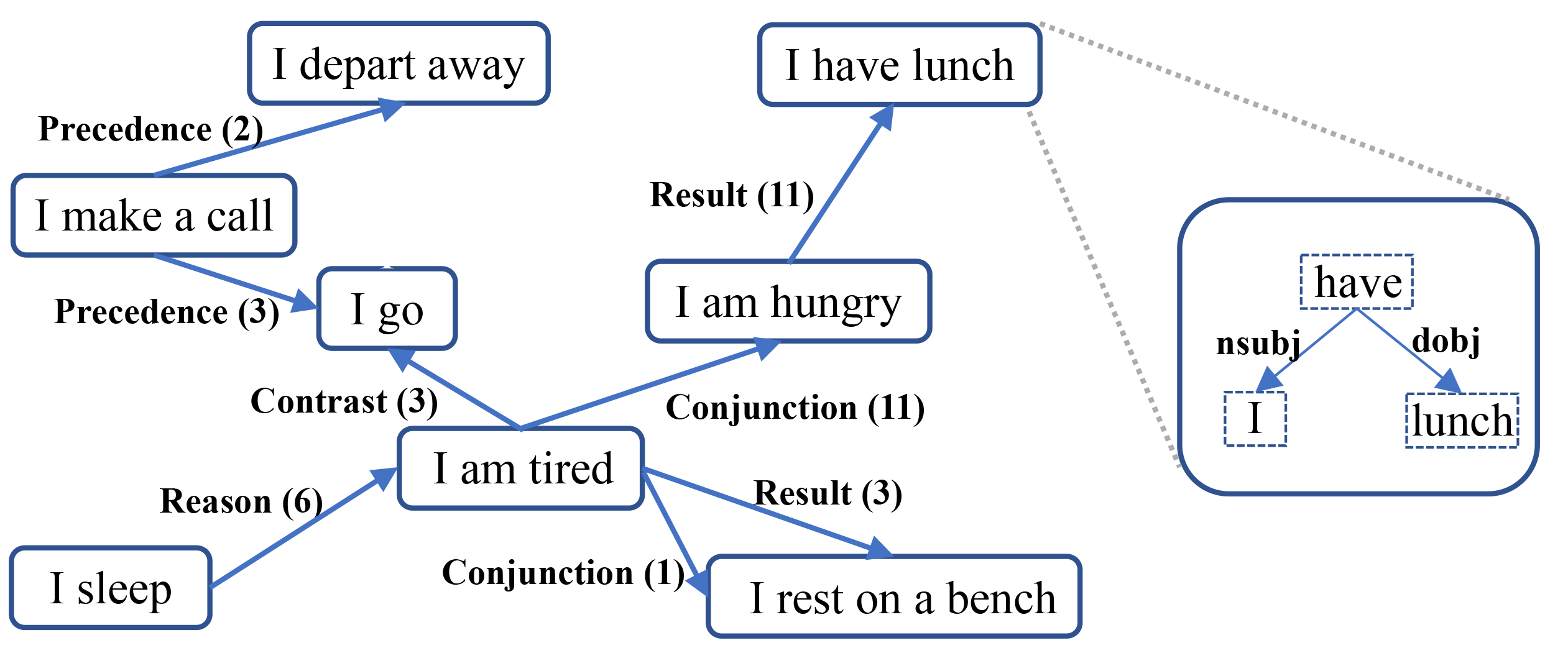}

\caption{ASER Demonstration. Eventualities are connected with weighted directed edges. Each eventuality is a dependency graph.}

\label{fig:ASER-demo}
% \vspace{-0.1in}
\end{figure}

For eventualities, traditional extraction approaches used in natural language processing based on FrameNet~\cite{framenet} or ACE~\cite{NIST05} first define complex structures of events by enumerating triggers with senses and arguments with roles. They then learn from limited annotated examples and try to generalize to other text contents.
%However, \revisehm{as} there are ambiguity and variability in semantic meanings of words and it is very difficult to generalize to new contents.
However, detecting trigger senses and argument roles suffers from the ambiguity and variability of the semantic meanings of words.
For example, using the ACE training data, the current state-of-the-art system can only achieve about 40\% overall F1 score with 33 event types~\cite{li2013joint}.
Different from them, we use patterns to extract eventuality-centric knowledge based on dependency grammar since the English language's syntax is relatively fixed and consistent across domains and topics. 
Instead of defining complex triggers and role structures of events, we simply use syntactic patterns to extract all possible eventualities.
We do not distinguish between semantic senses or categories of particular triggers or arguments in eventualities but treat all extracted words with their dependency relations as hyperedge in a graph to define an eventuality as a primitive semantic unit in our knowledge graph.
%To keep it unambiguous, we also do not annotate types of state, activity, and events although inherently we can use their definitions to distinguish them.

% and perform inference over them to identify interesting and useful patterns.
%An example of the structure of ASER is shown in Figure~\ref{fig:ASER-example} (\reviseyq{Example needed.})

For eventuality relations, we use the definition linguistic shallow discourse relations used in Penn Discourse Treebank (PDTB) \cite{prasad2007penn}. In PDTB, the relations are defined between two sentences or clauses. 
% The relations are difficult to determine when the sentences or clauses are long and complex.
% In ASER, we focus on simple and semantically complete patterns indicating eventualities.
Simplified from PDTB, we focus on relations between two eventualities, which are defined with simple but semantically complete patterns.
Moreover, as shown in PDTB, some connectives, e.g., `and' and `but', are less ambiguous than others, e.g., `while'.
Thus, we use less ambiguous connectives as seed connectives to find initial relations and then bootstrap the eventuality relation extraction using large corpora.
% \revisehm{Based on the extracted ASER, we also proposed a core version of ASER, which filter out eventualities and relations that appear only once and thus can have better accuracy.}
Although relations are extracted based on linguistic knowledge, we will show that they have correlations with previously defined commonsense knowledge in ConceptNet~\cite{liu2004conceptnet}.

%There exist several eventuality-related (or simply verb-related) knowledge bases, but they are inadequate for capturing the richness and complexity of eventualities and their relations. 
%There are also some existing knowledge bases about events (or just verb). However, they are still inadequate to model world's eventualities and their relations.

In total, ASER contains 194 million unique eventualities.
After bootstrapping, ASER contains 64 million edges among eventualities.
One example of ASER is shown in Figure~\ref{fig:ASER-demo}.
Table~\ref{tab:size_comparison} provides a size comparison between ASER and existing eventuality-related (or simply verb-centric) knowledge bases.
Essentially, they are not large enough as modern knowledge graph and inadequate for capturing the richness and complexity of eventualities and their relations.
FrameNet~\cite{framenet} is considered the earliest knowledge base defining events and their relations. It provides annotations about relations among about 1,000 human defined eventuality frames, which contain 27,691 eventualities. 
However, given the fine-grained definition of frames, the scale of the annotations is limited.
ACE~\cite{NIST05} (and its follow-up evaluation TAC-KBP~\cite{aguilar2014comparison}) reduces the number of event types and annotates more examples in each of event types.
PropBank~\cite{palmer2005proposition} and NomBank~\cite{meyers2004nombank} build frames over syntactic parse trees, and focus on annotating popular verbs and nouns.
TimeBank focuses only on temporal relations between verbs~\cite{pustejovsky2003timebank}.
While the aforementioned knowledge bases are annotated by domain experts, 
ConceptNet\footnote{Following the original definition, we only select the four relations (`HasPrerequisite', `HasFirstSubevent', `HasSubEvent', and `HasLastSubEvent') that involve eventualities.}~\cite{liu2004conceptnet}, Event2Mind~\cite{Event2Mind}, ProPora~\cite{proparNaacl2018}, and ATOMIC~\cite{Maarten2019Atomic} leveraged crowdsourcing platforms or the general public to annotate commonsense knowledge about eventualities, in particular the relations among them.
Furthermore, KnowlyWood~\cite{TandonMDW15KnowlyWood} uses semantic parsing to extract activities (verb+object) from movie/TV scenes and novels to build four types of relations (parent, previous, next, similarity) between activities using inference rules.
Compared with all these eventuality-related KGs, ASER is larger by one or more orders of magnitude in terms of the numbers of eventualities\footnote{Some of the eventualities are not connected with others, but the frequency of an eventuality is also valuable for downstream tasks. One example is the coreference resolution task. Given one sentence `The dog is chasing the cat, it barks loudly', we can correctly resolve `it' to `dog' rather than `cat' because `dog barks' appears 12,247 times in ASER, while `cat barks' never appears. This is usually called selectional preference~\cite{wilks1975preferential}, which has recently been evaluated in a larger scale in \cite{zhang2019sp-10k}. ASER naturally reflects human's selectional preference for many kinds of syntactic patterns.}
and relations it contains.

\begin{table}[t]
\footnotesize
    \centering
    \caption{Size comparison of ASER and existing eventuality-related resources. \# Eventuality, \# Relation, and \# R types are the number of eventualities, relations between these eventualities, and relation types. For KGs containing knowledge about both entity and eventualities, we report the statistics about the eventualities subset. ASER (core) filters out eventualities that appear only once and thus has better accuracy while ASER (full) can cover more knowledge.}
    {
    \begin{tabular}{l|c|c|c}
    \toprule
         & \# Eventuality & \# Relation & \# R Types \\
         \midrule
         FrameNet~\cite{framenet} & 27,691 & 1,709 & 7 \\
         ACE~\cite{aguilar2014comparison} & 3,290 & 0 & 0 \\
         PropBank~\cite{palmer2005proposition} &  112,917 & 0 & 0 \\ 
         NomBank~\cite{meyers2004nombank}  & 114,576 & 0 & 0 \\ 
         TimeBank~\cite{pustejovsky2003timebank} & 7,571 & 8,242 & 1 \\ % They record very detailed temporal relations between events 
        %*  Narrative & & & \\ % didn't find a widely used dataset
        %  Discourse Parsing & & & \\ can we treat each sentence as an event? it is a little wired.
         ConceptNet~\cite{liu2004conceptnet} & 74,989 & 116,097 & 4\\
         Event2Mind~\cite{Event2Mind} & 24,716 & 57,097 & 3\\
         ProPora~\cite{proparNaacl2018} & 2,406 & 16,269 & 1 \\
         ATOMIC~\cite{Maarten2019Atomic} & 309,515 & 877,108 & 9 \\
         Knowlywood~\cite{TandonMDW15KnowlyWood} & 964,758 & 2,644,415 & 4 \\ %only has temporal (prev: 761612, next: 757525), similarity (hassimilar: 342602) and isA (hashypernymy: 782676)
           \midrule
         ASER (core) & 27,565,673 & 10,361,178 & 15\\
         ASER (full) & 194,000,677 &  64,351,959 & 15\\
    \bottomrule
    \end{tabular}
    % \vspace{-0.2in}
    }
    \label{tab:size_comparison}
\end{table}

In summary, our contributions are as follows.

$\bullet$ {\bf Definition of ASER.} We define a brand new KG where the primitive units of semantics are eventualities. We organize our KG as a relational graph of hyperedges. 
Each eventuality instance is a hyperedge connecting several vertices, which are words. A relation between two eventualities in our KG represents one of the 14 relation types defined in PDTB~\cite{prasad2007penn} or a co-occurrence relation. 
%A relation is built between two eventualities to establish a certain relationship among the 14 relation \revisehm{type}s defined in PDTB~\cite{prasad2007penn} and co-occurrence.

$\bullet$ {\bf Scalable Extraction of ASER.} We perform eventuality extraction over large-scale corpora. We designed several high-quality patterns based on dependency parsing results and extract all eventualities that match these patterns. 
We use unambiguous connectives obtained from PDTB to find seed relations among eventualities. 
Then we leverage a neural bootstrapping framework to extract more relations from the unstructured textual data.

$\bullet$ {\bf Inference over ASER.} We also provide several ways of inference over ASER.
We show that both eventuality and relation retrieval over one-hop or multi-hop relations can be modeled as conditional probability inference problems.
% In fact, we can provide many conditional probabilities to show different kinds of semantics.

$\bullet$ {\bf Evaluation and Applications of ASER.} We conduct both intrinsic and extrinsic evaluations to validate the quality and effectiveness of ASER. For intrinsic evaluation, we sample instances of extracted knowledge in ASER over iterations, and submitted them to the Amazon Mechanical Turk (AMT) for human workers to verify. 
We also study the correlation of knowledge in ASER and the widely accepted commonsense knowledge in ConceptNet~\cite{liu2004conceptnet}.
% Each of the facts is evaluated by four workers on the Turk and the agreement of annotation is carefully evaluated.
For extrinsic evaluation, we use the Winograd Schema Challenge~\cite{levesque2011Winograd} to test whether ASER can help understand human language.
The results of both evaluations show that ASER is a promising large-scale KG with great potentials.

The proposed ASER and supporting packages are available at: \url{https://github.com/HKUST-KnowComp/ASER}.

\section{Overview of ASER}

\begin{table}[t]
	\footnotesize
	\centering
	\caption{Selected eventuality patterns (`v' stands for normal verbs other than `be', `be' stands for `be' verbs, `n' stands for nouns, `a' stands for adjectives, and `p' stands for prepositions.), Code (to save space, we create a unique code for each pattern and will use that in the rest of this paper), and the corresponding examples. 
	} \label{tab:eventuality-pattern}	   
	{
		\begin{tabular}{p{0.23\textwidth}|c|p{0.14\textwidth}}
			\toprule 
			Pattern & Code & Example \\
			\midrule
			$n_1$-nsubj-$v_1$ &s-v& `The dog barks' \\
			% 				$v_1$-dobj-$n_1$ &v-o& `pay the bill' \\
			$n_1$-nsubj-$v_1$-dobj-$n_2$ &s-v-o& `I love you' \\
			$n_1$-nsubj-$v_1$-xcomp-$a$ &s-v-a& `He felt ill' \\
			%                 ($v_1$-iobj-$n_1$)-dobj-$n_2$ &v-o-o& `give me the book' \\
			$n_1$-nsubj-($v_1$-iobj-$n_2$)-dobj-$n_3$ &s-v-o-o& `You give me the book'\\
			$n_1$-nsubj-$a_1$-cop-$be$ &s-be-a& `The dog is cute' \\
			% $n_1$-nsubj-$n_2$-cop-$be$ &s-be-o& `He is a boy' \\
			$n_1$-nsubj-$v_1$-xcomp-$a_1$-cop-$be$ &s-v-be-a& `I want to be slim'\\
			$n_1$-nsubj-$v_1$-xcomp-$n_2$-cop-$be$ &s-v-be-o& `I want to be a hero'\\
			$n_1$-nsubj-$v_1$-xcomp-$v_2$-dobj-$n_2$ &s-v-v-o&`I want to eat the apple' \\
			$n_1$-nsubj-$v_1$-xcomp-$v_2$ &s-v-v&`I want to go' \\
			($n_1$-nsubj-$a_1$-cop-$be$)-nmod-$n_2$-case-$p_1$ &s-be-a-p-o& `It's cheap for the quality' \\
			%                 $v_1$-nmod-$n_1$-case-$p_1$ &v-p-o& `Walk into the room'\\
			$n_1$-nsubj-$v_1$-nmod-$n_2$-case-$p_1$ &s-v-p-o& `He walks into the room'\\
			($n_1$-nsubj-$v_1$-dobj-$n_2$)-nmod-$n_3$-case-$p_1$ &s-v-o-p-o& `He plays soccer with me'\\
			$n_1$-nsubjpass-$v_1$ &spass-v& `The bill is paid'\\
			$n_1$-nsubjpass-$v_1$-nmod-$n_2$-case-$p_1$ &spass-v-p-o& `The bill is paid by me'\\
			%                 \midrule
			% 				$n_1$-nsubj-$v_1$-ccomp-($n_2$-nsubj-$v_2$-dobj-$n_3$) &s-v-(s-v-o)& `I think I love you'  & 0 & 0 & 0 & 0\\
			% 				$n_1$-nsubj-$v_1$-ccomp-($n_2$-nsubj-$a_1$-cop-$v_2$) &s-v-(s-v-a)& `I feel it is expensive'  & 0 & 0 & 0 & 0\\
			\bottomrule
		\end{tabular}
		% 			\vspace{-0.1in}
	}
\end{table}

%We first introduce the definitions and organization of our KG. 
%\subsection{Knowledge Representation and Notations}\label{sec:kg_notation}

%Different from most popular KGs, such as Freebase, which use triplets to denote entities and their relations, we focus on eventualities.

Each eventuality in ASER is represented by a set of words, where the number of words varies from one eventuality to another.
Thus, we cannot use a traditional graph representation such as triplets to represent knowledge in ASER. We devise the formal definition of our ASER KG as below.

\begin{definition}
{\bf ASER KG} is a hybrid graph $\HM$ of eventualities $E$'s. Each {\bf eventuality} $E$ is a hyperedge linking to a set of vertices $v$'s.
Each {vertex} $v$ is a {\bf word} in the vocabulary.
We define $v\in \VM$ in the vertex set and $E \in \EM$ in the hyperedge set.
$\EM \subseteq \PM(\VM)\setminus \{\emptyset\}$ is a subset of the power set of $\VM$.
We also define a {\bf relation} $R_{i,j}\in \RM$ between two eventualities $E_i$ and $E_j$, where $\RM$ is the relation set.
Each relation has a {\bf type} $T\in \TM$ where $\TM$ is the type set.
Overall, we have ASER KG $\HM=\{\VM, \EM, \RM, \TM\}$.
\end{definition}
ASER KG is a hybrid graph combining a hypergraph $\{\VM, \EM\}$ where each hyperedge is constructed over vertices, and a traditional graph $\{\EM, \RM\}$ where each edge is built among eventualities.
For example, $E_1$=\texttt{(I, am, hungry)} and $E_2$=\texttt{(I, eat, anything)} are eventualities, where we omit the internal dependency structures for brevity.
They have a relation $R_{1,2}$=\texttt{Result}, where \texttt{Result} is the relation type.
% \revisehm{In addition, we can also construct a bipartite graph based on the hypergraph $\{\VM, \EM\}$ where an edge can be build between a word and an eventuality.
% However, by nature $\{\VM, \EM\}$ is not a bipartite graph because we do not know eventualities in a priori until we construct them based on words.}

% \todo[inline]{Double check `s-be-o' pattern.}
        
\subsection{Eventuality}\label{sec:eventuality-pattern}

Different from named entities or concepts, which are noun phrases, eventualities are usually expressed as verb phrases, which are more complicated in structure. 
Our definition of eventualities is built upon the following two assumptions:
(1) syntactic patterns of English are relatively fixed and consistent; (2) the eventuality's semantic meaning is determined by the words it contains.
To avoid the extracted eventualities being too sparse, we use words fitting certain patterns rather than a whole sentence to represent an eventuality.
In addition, to make sure the extracted eventualities have complete semantics, 
we retain all necessary words extracted by patterns rather than those simple verbs or verb-object pairs in sentences.
% \revisehm{In addition, considering that eventuality consists of activity, state,to make sure the extracted eventualities have complete semantics}
The selected patterns are shown in Table~\ref{tab:eventuality-pattern}.
% The fixed patterns to extract eventualities are simple and semantically complete verb phrases.
For example, for the eventuality \texttt{(dog, bark)}, we have a relation \texttt{nsubj} between the two words to indicate that there is a subject-of-a-verb relation in between.
We now formally define an eventuality as follows.
\begin{definition}
An eventuality $E_i$ is a hyperedge linking multiple words $\{v_{i,1}, \ldots, v_{i,N_i} \}$, where $N_i$ is the number of words in eventuality $E_i$. Here, $v_{i,1}, \ldots, v_{i,N_i}\in \VM$ are all in the vocabulary. A pair of words in $E_i$ $(v_{i,j},v_{i,k})$ may follow a syntactic relation $e_{i,j,k}$.
\end{definition}

%Here we use $w_{i,j}$ in eventuality definition as we use it to denote different words while we use $v_i$ to denote unique words in the vocabulary.

We use patterns from dependency parsing to extract eventualities $E$'s from unstructured large-scale corpora.
Here $e_{i,j,k}$ is one of the relations that dependency parsing may return.
% Thus, we only consider simple verb phrase rather than those contain \revisehm{clauses}.
% In this sense, we are extracting more commonsense knowledge as events in OMCS.
Although in this way the recall is sacrificed, our patterns are of high precision and we use very large corpora to extract as many eventualities as possible. 
This strategy is also shared with many other modern KGs~\cite{knowitall,BankoCSBE07,NELL,wu2011taxonomy}.

\begin{table}[t]
			\centering
			\footnotesize
			\caption{Eventuality relation types between two eventualities $E_1$ and $E_2$ and explanations.}	\label{tab:relation-def}
			{\
			\begin{tabular}{p{0.2\textwidth}|p{0.24\textwidth}}
				\toprule 
				Relation & Explanation  \\
				\midrule
				$<$$E_1$, `Precedence', $E_2$$>$ & $E_1$ happens before $E_2$. \\
				$<$$E_1$, `Succession', $E_2$$>$ & $E_1$ happens after $E_2$. \\
				$<$$E_1$, `Synchronous', $E_2$$>$ & $E_1$ happens at the same time as $E_2$. \\ \midrule
				$<$$E_1$, `Reason', $E_2$$>$ & $E_1$ happens because $E_2$ happens. \\
				$<$$E_1$, `Result', $E_2$$>$ & If $E_1$ happens, it will result in the happening of $E_2$. \\
				$<$$E_1$, `Condition', $E_2$$>$ & Only when $E_2$ happens, $E_1$ can happen. \\ \midrule
				$<$$E_1$, `Contrast', $E_2$$>$ & $E_1$ and $E_2$ share a predicate or property and have significant difference on that property. \\
				$<$$E_1$, `Concession', $E_2$$>$ & $E_1$ should result in the happening of $E_3$, but $E_2$ indicates the opposite of $E_3$ happens. \\ \midrule
				$<$$E_1$, `Conjunction', $E_2$$>$ & $E_1$ and $E_2$ both happen. \\
				$<$$E_1$, `Instantiation, $E_2$$>$ & $E_2$ is a more detailed description of $E_1$. \\
				$<$$E_1$, `Restatement', $E_2$$>$ & $E_2$ restates the semantics meaning of $E_1$.\\ 
				$<$$E_1$, `Alternative', $E_2$$>$ & $E_1$ and $E_2$ are alternative situations of each other. \\
				$<$$E_1$, `ChosenAlternative', $E_2$$>$ & $E_1$ and $E_2$ are alternative situations of each other, but the subject prefers $E_1$. \\
				$<$$E_1$, `Exception', $E_2$$>$ & $E_2$ is an exception of $E_1$. \\ \midrule
				$<$$E_1$, `Co-Occurrence', $E_2$$>$ & $E_1$ and $E_2$ appear in the same sentence. \\
				\bottomrule
			\end{tabular}
			}
        \end{table}
        
\subsection{Eventuality Relation}

For relations among eventualities, as introduced in Section~\ref{sec:introduction}, we follow PDTB's~\cite{prasad2007penn} definition of relations between sentences or clauses but simplify them to eventualities.
Following the CoNLL 2015 discourse parsing shared task~\cite{xue2015conll}, we select 14 discourse relation types and an additional co-occurrence relation to build our knowledge graph.

\begin{definition}\label{def:relations}
A relation $R$ between a pair of eventualities $E_1$ and $E_2$ has one of the following types $T\in\TM$ and all types can be grouped into five categories: 
{\bf Temporal} (including {Precedence}, {Succession}, and {Synchronous}),
{\bf Contingency} (including {Reason}, {Result}, and {Condition}),
{\bf Comparison} (including {Contrast} and {Concession}),
{\bf Expansion} (including {Conjunction}, {Instantiation}, {Restatement}, {Alternative}, {ChosenAlternative}, and {Exception}), and
{\bf Co-Occurrence}.
The detailed definitions of these relation types are shown in Table~\ref{tab:relation-def}.
The weight of $R$ is defined by the number of tuple $<$$E_1$, $R$, $E_2$$>$ appears in the whole corpora.
\end{definition}

%As one of the most important relation among eventualities, temporal relations are used to record the order of eventualities happening in our daily life. The most important temporal relations .

% \begin{definition}
% {\bf Contingency.}
% Contingency relations are used to model the logical relations between two eventualities. These relations are crucial for answering `why' questions. It includes following three sub-categories: \texttt{Reason}, \texttt{Result}, and \texttt{Condition}.
% \end{definition}

% \begin{definition}
% {\bf Comparison.}
% Comparison relations indicate whether two eventualities have similar meaning or belong to the same category or whether they are different from others.
% It includes \texttt{Contrast} and \texttt{Concession} sub-categories.
% \end{definition}

% \begin{definition}
% {\bf Expansion.} 
% Expansion indicates the expansion discourse and move its narrative or exposition forward. It includes six sub-categories: \texttt{Conjunction}, \texttt{Instantiation}, \texttt{Restatement}, \texttt{Alternative}, \texttt{ChosenAlternative}, and \texttt{Exception}.
% \end{definition}

% \begin{definition}
% {\bf Co-Occurrence.}
% A \texttt{co-occurrence} relation between eventualities means that two eventualities frequently appear in the same sentence.
% \end{definition}

\subsection{KG Storage}
All eventualities in ASER are small-dependency graphs, where vertices are the words and edges are the internal dependency relations between these words. 
We store the information about eventualities and relations among them separately in two tables with a SQL database.
In the eventuality table, we record information about event ids, all the words, dependencies edges between words, and frequencies.
% Besides that, to support partial matches of eventualities, we also record information about lemma words and verbs contained in the eventualities.
In the relation table, we record ids of head and tail eventualities and relations between them.

\section{Knowledge Extraction}\label{sec:framework}
In this section, we introduce the knowledge extraction methodologies for building ASER.

\begin{figure}
\centering

\includegraphics[width=\linewidth]{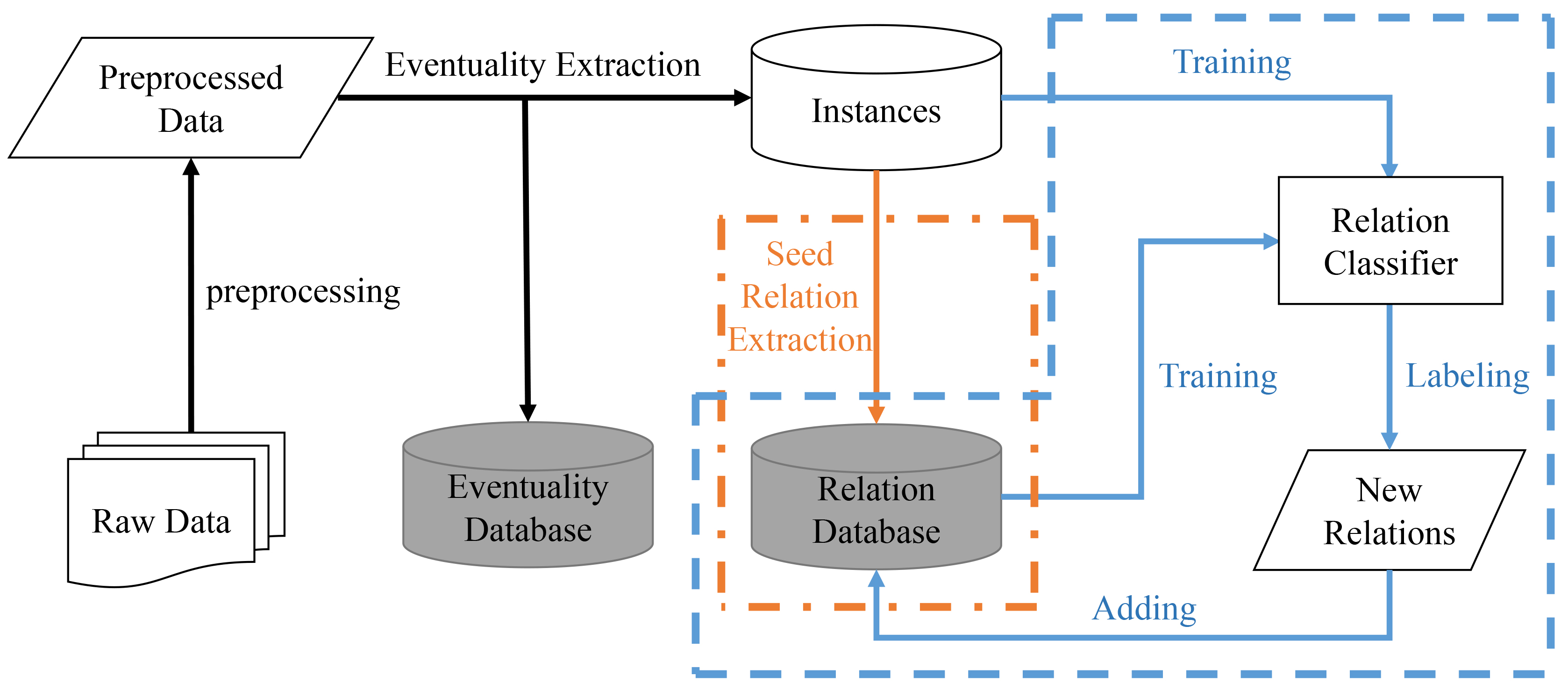}
\caption{ASER extraction framework. The seed relation selection and the bootstrapping process are shown in the orange dash-dotted and blue dashed box respectively. Two gray databases are the resulted ASER.}
\label{fig:framework}
\end{figure}

%Cane: you may want to add a section here to describe the formal definitions of eventuality and relations. The "Framework" section can then focus on the methods for extracting/learning them. 

\subsection{System Overview}\label{sec:system-overview}
We first introduce the overall framework of our knowledge extraction system. The framework is shown in Figure~\ref{fig:framework}.
After textual data collection, we first preprocess the texts with the dependency parser.
Then we perform eventuality extracting using pattern matching.
For each sentence, if we find more than two eventualities, we first group these eventualities into pairs. For each pair, we generate one training instance, where each training instance contains two eventualities and their original sentence.
After that, we extract seed relations from these training instances based on the less ambiguous connectives obtained from PDTB~\cite{prasad2007penn}. 
Finally, a bootstrapping process is conducted to learn more relations and train the new classifier repeatedly.
In the following sub-sections, we will introduce each part of the system separately.

\subsection{Corpora}

To make sure the broad coverage of ASER, we select corpora from different resources (reviews, news, forums, social media, movie subtitles, e-books) as the raw data. The details of these datasets are as follows.

$\bullet$ Yelp: Yelp is a social media platform where users can write reviews for businesses, e.g., restaurants, hotels. The latest release of the Yelp dataset\footnote{\url{https://www.yelp.com/dataset/challenge}} contains over five million reviews. 

$\bullet$ New York Times (NYT): The NYT~\cite{nyt} corpus contains over 1.8 million news articles from the NYT throughout 20 years (1987 - 2007).

        \begin{table}[t]
        \footnotesize
			\centering
			\caption{Statistics of used corpora. (M means millions.)} \label{tab:corpora-statistics}		
			{
			\begin{tabular}{c|c|c|c|c}
				\toprule 
				Name   & \# Sentences & \# Tokens & \# Instances & \# Eventualities \\ % & Corpus Size & Category
				\midrule
				Yelp             & 48.9M & 758.3M & 54.2M &  20.5M \\ %& 2.5G  & Reviews
                NYT               & 56.8M & 1,196.9M & 41.6M & 23.9M \\ % & 3G  & News 
                Wiki          & 105.1M & 2,347.3M & 38.9M & 38.4M \\ % & 13G & Knowledge  
                Reddit          & 235.9M & 3,373.2M & 185.7M & 82.6M \\ % & 21G  & Forum  
                Subtitles & 445.0M & 3,164.1M & 137.6M & 27.0M \\ % & 13G  & Movie Scripts
                E-books       & 27.6M & 618.6M & 22.1M & 11.1M \\ %  & 3G  & Stories 
                \midrule
                Overall          & 919.2M & 11,458.4M & 480.1M & 194.0M \\ % & 55.5G  & -    
				\bottomrule
			\end{tabular}
			}
        \end{table}

$\bullet$ Wiki: Wikipedia is one of the largest free knowledge dataset. To build ASER, we select the English version of Wikipedia\footnote{\url{https://dumps.wikimedia.org/enwiki/}}.

$\bullet$ Reddit: Reddit is one of the largest online forums. In this work, we select the anonymized post records\footnote{\url{https://www.reddit.com/r/datasets/comments/3bxlg7}} over one period month.  

$\bullet$ Movie Subtitles: The movie subtitles corpus was collected by~\cite{lison2016opensubtitles2016} and we select the English subset, which contains subtitles for more than 310K movies.

$\bullet$ E-books: The last resource we include is the free English electronic books from Project Gutenberg\footnote{\url{https://www.gutenberg.org/}}.
% 5860 books

We merge these resources as a whole to perform knowledge extraction. The statistics of different corpora are shown in Table~\ref{tab:corpora-statistics}.

\begin{algorithm}[t]\caption{Eventuality Extraction with One Pattern $P_i$}\label{algorithm:eventuality-extraction}
\footnotesize
            \begin{flushleft}
			\textbf{INPUT:} Parsed dependency graph $D$, center verb $v$. Positive dependency edges $P_i^p$, optional edges $P_i^o$, and negative edges $P_i^n$.
            \textbf{OUTPUT:} Extracted eventuality $E$.
             \end{flushleft}
            \begin{algorithmic}[1]
            \State Initialize eventuality $E$.
            \For{Each connection $d$ (a relation and the associated word) in positive dependency edges $P_i^p$}  
                \If{Find $d$ in $D$}
                    \State Append $d$ in $E$.
                \Else
                	\State Return NULL.
                \EndIf 
            \EndFor
            \For{Each connection $d$ in optional dependency edges $P_i^o$}  
                \If{Find $d$ in $D$}
                      \State Append $d$ in $E$.
                \EndIf 
            \EndFor
            \For{Each connection $d$ in negative dependency edges $P_i^n$}  
               \If{Find $d$ in $D$}
                   \State Return NULL.
               \EndIf 
            \EndFor
            \State Return $E$
            \end{algorithmic}
        \end{algorithm}

\subsection{Preprocessing and Eventuality Extraction}\label{sec:eventuality-extraction}

For each sentence $s$, we first parse it with the Stanford Dependency Parser\footnote{\url{https://nlp.stanford.edu/software/stanford-dependencies.html}}. 
We then filter out all the sentences that contain clauses.
As each sentence may contain multiple eventualities and verbs are the centers of them, we first extract all verbs.
To make sure that all the extracted eventualities are semantically complete without being too complicated, we design 14 patterns to extract the eventualities via pattern matching. Each of the patterns contains three kinds of dependency edges: positive dependency edges, optional dependency edges, and negative dependency edges.
All positives edges are shown in Table~\ref{tab:eventuality-pattern}.
Six more dependency relations (\texttt{advmod}, \texttt{amod}, \texttt{nummod}, \texttt{aux}, \texttt{compound}, and \texttt{neg}) are optional dependency edges that can associate with any of the selected patterns.
We omit all optional edges in the table because they are the same for all patterns.
All other dependency edges are considered are negative dependency edges, which are designed to make sure all the extracted eventualities are semantically complete and all the patterns are exclusive with each other.
Take sentence `I have a book' as an example, we will only select $<$`I', `have', `book'$>$ rather than $<$`I', `have'$>$ as the valid eventuality, because `have'-dobj-`book' is a negative dependency edge for pattern `s-v'. 
For each verb $v$ and each pattern, we first put it in the position of $v_1$ and then try to find all the positive dependency edges. If we can find all the positive dependency edges around the center verb we consider it as one potential valid eventuality and then add all the words connected via those optional dependency edges. In the end, we will check if any negative dependency edge can be found in the dependency graph. If not, we will keep it as one valid eventuality. Otherwise, we will disqualify it. 
The pseudo-code of our extraction algorithm is shown in Algorithm~\ref{algorithm:eventuality-extraction}.
The time complexity of eventuality extraction is $\mathcal{O}(|S| \cdot |D| \cdot |v|)$ where $|S|$ is the number of sentences, $|D|$ is the average number of dependency edges in a dependency parse tree, and $|v|$ is the average number of verbs in a sentence.

%         \begin{table}[t]
% 			\centering
% 			\caption{Statistics of used corpora data. (For \# eventualities, we mean the number of unique eventualities.)} \label{tab:corpora-statistics}		
% 			{\small
% 			\begin{tabular}{c|c|c|c|c|c}
% 				\toprule 
% 				Name & Category  & \# Sentences & \# Tokens & \# Examples & \# Eventualities \\ % & Corpus Size
% 				\midrule
% 				Yelp      & Reviews        & 48,917,647 & 758,346,258 & 54,190,294 &  20,478,418 \\ %& 2.5G 
%                 NYT       & News          & 56,806,032 & 1,196,918,667 & 41,614,823 & 23,877,703 \\ % & 3G 
%                 Wiki      & Knowledge      & 105,075,347 & 2,347,278,848 & 38,938,199 & 38,375,106 \\ % & 13G
%                 Reddit    & Forum         & 235,887,505 & 3,373,164,871 & 185,667,502 & 82,643,183 \\ % & 21G 
%                 Subtitles & Movie Scripts & 444,963,752 & 3,164,072,668 & 137,610,354 & 27,013,622 \\ % & 13G 
%                 E-books   & Stories       & 27,593,543 & 618,631,540 & 22,064,170 & 11,085,848 \\ %  & 3G
%                 \midrule
%                 Overall   & -             & 919,243,826 & 11,458,412,852 & 480,085,342 & 194,000,680 \\ % & 55.5G
% 				\bottomrule
% 			\end{tabular}
% 			}
%         \end{table}

        \begin{table}[t]
        \footnotesize
			\centering
			\caption{Selected seed connectives. Here relations are directed relation from $E_1$ to $E_2$. Each relation can have multiple seed connectives, where the corresponding connectives are highlighted as boldface.}\label{tab:seed-pattern}		
			{
		    	\begin{tabular}{p{0.12\textwidth}|p{0.32\textwidth}}
            
				\toprule 
				Relation Type & Seed Patterns  \\
				\midrule
				Precedence      & $E_1$ {\bf before} $E_2$; $E_1$ {\bf, then} $E_2$; $E_1$ {\bf till} $E_2$; $E_1$ {\bf until} $E_2$ \\\hline
				Succession     & $E_1$ {\bf after} $E_2$; $E_1$ {\bf once} $E_2$\\\hline
                Synchronous  & $E_1$, {\bf meanwhile} $E_2$; $E_1$ {\bf meantime} $E_2$; $E_1$, {\bf at the same time} $E_2$\\
                \midrule
                Reason     & $E_1$, {\bf because} $E_2$ \\\hline
                Result      & $E_1$, {\bf so} $E_2$; $E_1$, {\bf thus} $E_2$; $E_1$, {\bf therefore} $E_2$; $E_1$, {\bf so that} $E_2$\\\hline
                % ['accordingly'], ['as', 'a', 'result'], ['so'], ['so', 'that'],['thus'], ['therefore']
%                 $E_1$, thereby $E_2$  $E_1$, hence $E_2$
                Condition    &  $E_1$, {\bf if} $E_2$; $E_1$, {\bf as long as} $E_2$ \\
                \midrule
                Contrast    & $E_1$, {\bf but} $E_2$; $E_1$, {\bf however} $E_2$; $E_1$, {\bf, by contrast} $E_2$; $E_1$, {\bf, in contrast} $E_2$; $E_1$, {\bf, on the other hand,} $E_2$; $E_1$, {\bf , on the contrary,} $E_2$\\\hline
%                 nonetheless, on the contrary, on the other hand
				Concession     & $E_1$, {\bf although} $E_2$ \\
                \midrule
                Conjunction  & $E_1$ {\bf and} $E_2$;  $E_1$, {\bf also} $E_2$;\\\hline
%                 likewise moreover plus similarly
                Instantiation & $E_1$, {\bf for example} $E_2$; $E_1$, {\bf for instance} $E_2$ \\\hline
                Restatement     & $E_1$, {\bf in other words} $E_2$ \\\hline
                Alternative   & $E_1$ {\bf or} $E_2$; $E_1$, {\bf unless} $E_2$; $E_1$, {\bf as an alternative} $E_2$; $E_1$, {\bf otherwise} $E_2$\\\hline
                ChosenAlternative     & $E_1$, $E_2$ {\bf instead} \\\hline
                Exception & $E_1$, {\bf except} $E_2$ \\

				\bottomrule
			\end{tabular}
			}
           \end{table}

\subsection{Eventuality Relation Extraction}
           
For each training instance, we use a two-step approach to decide the relations between the two eventualities.

We first extract seed relations from the corpora by using the unambiguous connectives obtained from PDTB~\cite{prasad2007penn}.
According to PDTB's annotation manual, we found that some of the connectives are more unambiguous than the others.
For example, in the PDTB annotations, the connective `so that' is annotated 31 times and is only with the \texttt{Result} relation.
On the other hand, the connective `while' is annotated as \texttt{Conjunction} 39 times,  \texttt{Contrast} 111 times, \texttt{expectation} 79 times, and \texttt{Concession} 85 times. 
When we identify connectives like `while', we can not determine the relation between the two eventualities related to it.
Thus, we choose connectives that are less ambiguous, where more than 90\% annotations of each are indicating the same relation, to extract seed relations.
The selected connectives are listed in Table~\ref{tab:seed-pattern}.
Formally, we denote one informative connective word(s) and its corresponding relation type as $c$ and $T$. Given a training instance $x$=($E_1$, $E_2$, $s$), if we can find a connective $c$ such that $E_1$ and $E_2$ are connected by $c$ according to the dependency parse, we will select this instance as an instance for relation type $T$. 
% After annotating the less ambiguous relations, to make sure the high quality of our seed instances, for each relation $R$, we use global statistics to find event relations that are frequently annotated as seed relations to do further processing.

%we only select the most representative connective words $s_1, s_2, ...$ to create seed patterns, which means that in the original penn annotation guide~\cite{prasad2007penn}, more than 90\% of these selected words are indicating relation $R$. 

Since the seed relations extracted with selected connectives can only cover the limited number of the knowledge, we use a bootstrapping framework to incrementally extract more eventuality relations.
Bootstrapping~\cite{agichtein2000snowball} is a commonly used technique in information extraction.
Here we use a neural network based approach to bootstrap.
The general steps of bootstrapping are as follows.

$\bullet$ Step 1: Use the extracted seed training instances as the initial labeled training instances.

$\bullet$ Step 2: Train a classifier based on labeled training instances.

$\bullet$ Step 3: Use the classifier to predict relations of each training instance. If the prediction confidence of certain relation type $T$ is higher than the selected threshold, we will label this instance with $T$ and add it to the labeled training instances.
% \revisexin{We believe predictions in previous iterations are more confident so that we just annotate those instances without relations or whose new predicted probabilities are higher.} 
Then go to Step 2.

\begin{figure}
\centering
\includegraphics[width=\linewidth]{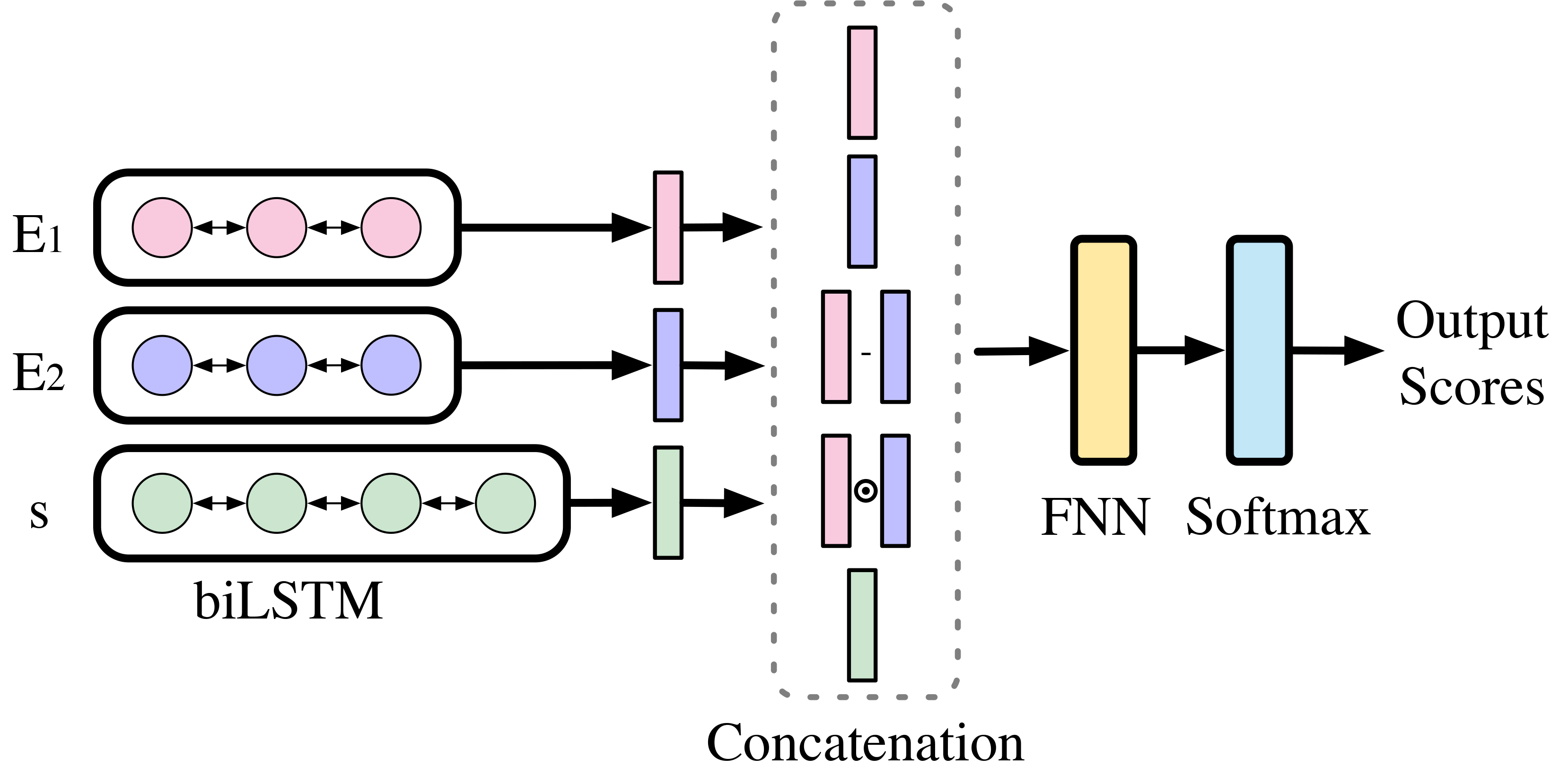}
\caption{The overview of the neural classifier. For each instance $x=(E_1,E_2,s)$, we first encode the information of two eventualities $E_1$, $E_2$ and the original sentence $s$ with three bidirectional LSTMs~\cite{hochreiter1997long} module
% \footnote{We also tried Transformer network used in BERT~\cite{devlin2018bert}, but in our experiments the attention emphasizes too much over the seed connectives and cannot effectively find new edges.}
and the output representations are $h_{E_1}$, $h_{E_2}$ and $h_{s}$ respectively.
We then concatenate $h_{E_1}$, $h_{E_2}$, $h_{E_1}-h_{E_2}$, $h_{E_1} \circ h_{E_2}$ and $h_{s}$ together, where $\circ$ indicates the element-wise multiplication, and feed them to a two-layer feed forward network.
In the end, we use a softmax function to generate scores for different relation types. 
}
\label{fig:bootstrapper}
\end{figure}
% \vspace{-0.5in}

The neural classifier architecture is shown in Figure~\ref{fig:bootstrapper}. In the training process, we randomly select labelled training instances as positive examples and unlabelled training instances as negative examples. The cross-entropy is used as the loss and the whole model is updated via Adam~\cite{kingma:adam}.
In the labeling process, for each training instance $x$, the classifier can predict a score for each relation type.
For any relation type, if the output score is larger than a threshold $\tau_k$, where $k$ is the number of bootstrapping iteration, we will label $x$ with that relation type.
To avoid error accumulation, we also use the annealing strategy to increase the threshold $\tau_k=\tau_0 + (1-\tau_0) / (1 + \exp{(-(k-K/2))})$, where $K$ is the total iteration number.
The complexities of both training and labeling processes in $k^{th}$ iteration are linear to the number of parameters in LSTM cell $|L|$, the number of training examples $|I_{{train}_k}|$, and the number of instances to predict $|I_{{predict}_k}|$ in $k^{th}$ iteration. So the overall complexity in $k^{th}$ iteration is $\mathcal{O}(|L| \cdot  (|I_{{train}_k}| + |I_{{predict}_k}|))$.

Used hyper-parameters and other implementation details are as follows:
For preprocessing, we first parse all raw corpora with the Stanford Dependency parser, which costs eight days with two 12-core Intel Xeon Gold 5118 CPUs. After that, We extract eventualities, build the training instance set, and extract seed relations, which costs two days with the same CPUs.
For bootstrapping, 
Adam optimizer \cite{kingma:adam} is used and the initial learning rate is 0.001. The batch size is 512. We use GloVe~\cite{DBLP:conf/emnlp/PenningtonSM14} as the pre-trained word embeddings. The dropout rate is 0.2 to prevent overfitting. The hidden sizes of LSTMs are 256 and the hidden size of the two-layer feed forward network with ReLU is 512. 
As relation types belonging to different categories could both exist in one training instance, in each bootstrapping iteration, four different classifiers are trained corresponding to four categories (\textbf{Temporal}, \textbf{Contingency}, \textbf{Comparison}, \textbf{Temporal}).
Each classifier predicts the types belong to that category or `None' of each instance. Therefore, classifiers do not influence each other so that they can be processed in parallel. Each iteration using ASER (core) takes around one hour with the same CPUs and four TITAN X GPUs. We spend around eight hours predicting ASER (full) with the learned classifier in the 10th iteration.

\section{Inference over ASER}~\label{sec:inference}

In this section, we provide two kinds of inferences (eventuality retrieval and relation retrieval) based on ASER. For each of them, inferences over both one-hop and multi-hops are provided. Complexities of these two retrieval algorithms are both $\mathcal{O}(A^k)$, where $A$ is the number of average adjacent eventualities per eventuality and $k$ is the number of hops. 
In this section, we show how to conduct these inferences over one-hop and two-hop as the demonstration.

\subsection{Eventuality Retrieval}
The eventuality retrieval inference is defined as follows. Given a head eventuality\footnote{ASER also supports the prediction of head eventualities given tail eventuality and relations. We omit it in this section for the clear presentation.} $E_h$ and a relation list $\LM$ = ($R_1, R_2, ..., R_k$), find related eventualities and their associated probabilities such that for each eventuality $E_t$ we can find a path, which contains all the relations in $\LM$ in order from $E_h$ to $E_t$.
\subsubsection{One-hop Inference}

For the one-hop inference, we assume the target relation is $R_1$. We then define the probability of any potential tail eventuality $E_t$ as:
\begin{equation}\label{eq:one-hop-eventuality-retrieval}
    P(E_t| E_h, R_1) = 
    \frac{f(E_h, R_1, E_t)}{\sum_{E_t^\prime,s.t.,(E_t,R_1)\in ASER}f(E_h, R_1, E_t^\prime)},
\end{equation}
where $f(E_h, R_1, E_t)$ is the relation weight, which is defined in Definition 3. If no eventuality is connected with $E_h$ via $R_1$, $P(E^\prime | E_h, R)$ will be 0 for any $E^\prime \in \EM$. 

\subsubsection{Two-hop Inference}
On top of Eq.~(\ref{eq:one-hop-eventuality-retrieval}), it is easy for us to define the probability of $E_t$ on two-hop setting. Assume the two relations are $R_1$ and $R_2$ in order. We can define the probability as follows:
\begin{equation}\label{eq:two-hop-eventuality-retrieval}
    P(E_t| E_h, R_1, R_2) = 
    \sum_{E_m\in \EM_m}P(E_m|E_h, R_1)P(E_t|E_m, R_2),
\end{equation}
where $\EM_m$ is the set of intermediate eventuality $E_m$ such that $(E_h,R_1,E_m)$ and $(E_m, R_2, E_t) \in ASER$.

\subsection{Relation Retrieval}

The relation retrieval inference is defined as follows. Given two eventualities $E_h$ and $E_t$, find all relation lists and their probabilities such that for each relation list $\LM$ = ($R_1, R_2, ..., R_k$), we can find a path from $E_h$ to $E_t$, which contains all the relations in $\LM$ in order.

\subsubsection{One-hop Inference}

Assuming that the path length is one, we define the probability of one relation $R$ exist from $E_h$ to $E_t$ as:
\begin{equation}\label{eq:one-hop-plausibility}
    P(R| E_h, E_t) = 
    \frac{f(E_h, R, E_t)}{\sum_{R^\prime \in \RM}f(E_h, R^\prime, E_t)},
\end{equation}
where $\RM$ is the relation set.
% After that, for each relation type $T \in \TM$, we can find the most possible relation via:
% \begin{equation}
%   R_{max}  = \argmax_{R^\prime \in \RM} P(R^\prime| E_h, E_t),
% \end{equation}
% where $P$ is the aforementioned plausibility scoring function and $\R$ is the relations set. 
% One thing worth mentioning is that to make sure the quality of extracted relation, ASER will return $R_{max}$, only when $P(R_{max}|E_h, E_t)$ is larger than 0.5 and will return `NULL' otherwise.

\begin{figure}[t!]
\centering
        \includegraphics[width=\linewidth]{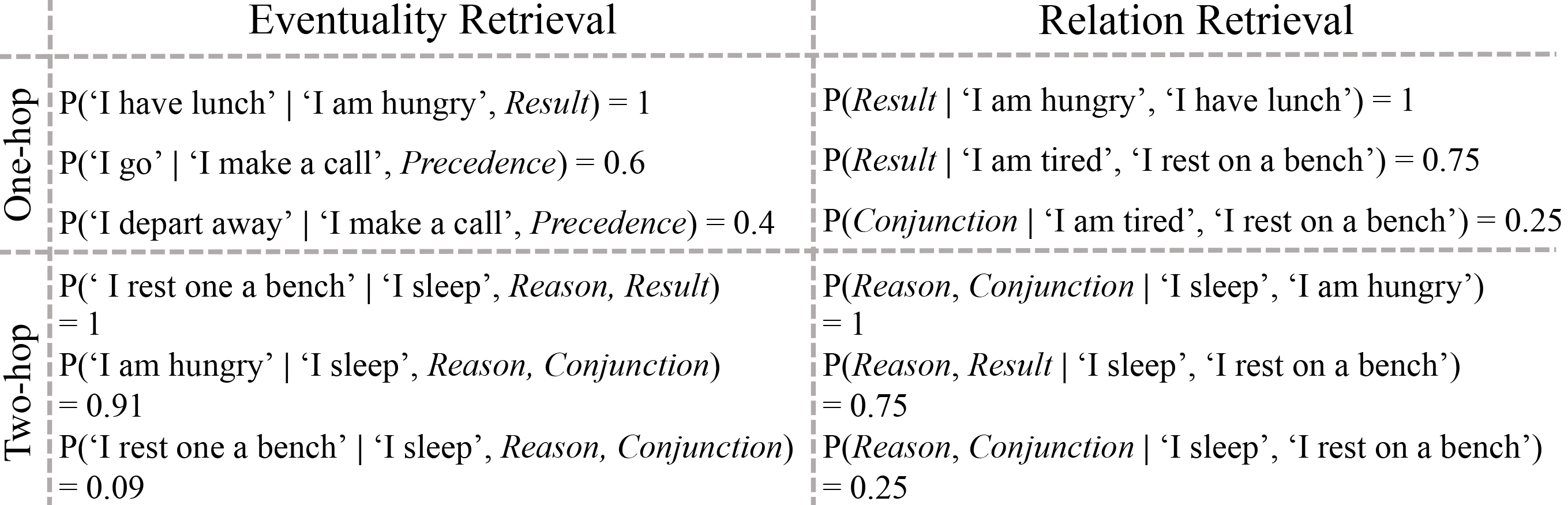}
        \caption{Examples of inference over ASER.}\label{fig:inference}
\end{figure}

\subsubsection{Two-hop Inference}
Similarly, given two eventualities $E_h$ and $E_t$, we define the probability of a two-hop connection ($R_1$, $R_2$) between them as follows:

\begin{align}
    P(R_1,R_2|E_h,E_t)  &=\sum_{E_m\in \EM_m}P(R_1,R_2,E_m|E_h,E_t) \nonumber\\
    &=\sum_{E_m\in \EM_m}P(R_1|E_h)P(E_m|R_1,E_h)P(R_2|E_m,E_t),
\end{align}
where $P(R|E_h)$ is the probability of relation $R$, given head eventuality $E_h$, and is defined as follows:
\begin{equation}\label{eq:relation-probability}
    P(R|E_h)=
    \frac{\sum_{E_t,s.t.,(E_t,R)\in ASER}f(E_h, R, E_t)}{\sum_{R^\prime\in \RM}\sum_{E_t,s.t.,(E_t,R)\in ASER}f(E_h, R^\prime, E_t)}.
\end{equation}
% We can then get the most possible relation pair via:
% \begin{equation}
%     (R_{1, max}, R_{2,max}) = \argmax_{R_1^\prime, R_2^\prime \in \R} P(E_h,R_1^\prime,R_2^\prime,E_t).
% \end{equation}
% Similar to the one-hop inference, here $P(E_h,R_{1.max},R_{2,max},E_t)$ is required to be large than 0.5 and we will return `NULL' otherwise.

% On top of that, if we have one eventuality $e$, we can find the eventuality $e^\prime$ connected with $e$ with two relations $r_1$ and $r_2$ in order by requiring $P_2(e_h,r_1,r_2,e_t)$ to be larger than a threshold $t_2$. Of course, the complexity of this search algorithm is $O(|N|^2)$, where $|N|$ is the number of neighbors of each eventuality.
% On the other way around, if we already have the head and tail eventualities $e_h$ and $e_t$, 

\subsection{Case Study}
        \begin{table}[t]
	\footnotesize
	\centering
	\caption{Statistics and annotations of the eventuality extraction. \# Eventuality and \# Unique means the total number and the unique number of extracted eventualities using corresponding patterns  (`M' stands for millions). \# Agreed means the number of agreed eventualities among five annotators. \# Valid means the number valid eventualities labeled by annotators. Accuracy (Acc.)=\# Valid/\# Agrees. The Overall accuracy is calculated based on the pattern distribution.}\label{tab:eventuality-statistics}	
	{
		\begin{tabular}{c|c|c|c|c|c}
			\toprule 
			Pattern  & \# Eventuality & \# Unique  & \# Agreed & \# Valid  & Acc. \\
			\midrule
			s-v         & 109.0M   & 22.1M    & 171 & 158 & 92.4\% \\
			s-v-o       & 129.0M   & 60.0M    & 181 & 173 & 95.6\% \\
			s-v-a       & 5.2M     & 2.1M     & 195 & 192 & 98.5\% \\
			s-v-o-o     & 3.5M     & 1.7M     & 194 & 187 & 96.4\% \\
			s-be-a      & 89.9M    & 29.0M    & 189 & 188 & 99.5\% \\
			s-v-be-a    & 1.2M     & 0.5M     & 190 & 187 & 98.4\% \\
			s-v-be-o    & 1.2M     & 0.7M     & 186 & 171 & 91.9\% \\
			s-v-v-o     & 12.4M    & 6.6M     & 193 & 185 & 95.9\% \\
			s-v-v       & 8.7M     & 2.7M     & 185 & 155 & 83.8\% \\
			s-be-a-p-o  & 13.2M    & 8.7M     & 189 & 185 & 97.9\% \\
			s-v-p-o     & 39.0M    & 23.5M    & 178 & 161 & 90.4\% \\
			s-v-o-p-o   & 27.2M    & 19.7M    & 181 & 167 & 92.2\% \\
			spass-v     & 15.1M    & 6.2M     & 177 & 155 & 87.6\% \\
			spass-v-p-o & 13.5M    & 10.3M    & 188 & 177 & 94.1\% \\
			\midrule
			Overall     & 468.1M   & 194.0M   & --- & --- & 94.5\% \\
			\bottomrule
		\end{tabular}
	}
\vspace{-0.1in}
\end{table}

In this section, we showcase several interesting inference examples with ASER in Figure~\ref{fig:inference}, which is conducted over the extracted sub-graph of ASER shown in Figure~\ref{fig:ASER-demo}.
By doing inference over eventuality retrieval, we can easily find out that `I am hungry' usually results in having lunch and the eventuality `I make a call' often happens before someone goes or departs. More interestingly, leveraging the two-hop inference, given the eventuality `I sleep', we can find out an eventuality `I rest on a bench' such that both of them are caused by the same reason, which is `I am tired' in this example.
On the other hand, we can also retrieve possible relations between eventualities. For example, we can know that `I am hungry' is most likely the reason for `I have lunch' rather than the other way around.
Similarly, over the 2-hop inference, we can find out that even though `I am hungry' has no direct relation with `I sleep', `I am hungry' often appears at the same time with `I am tired', which is one plausible reason for `I sleep'.

% \section{Evaluations}\label{sec:evaluation}
% In this section, we present human evaluation and extrinsic experiments to evaluate the quality of ASER. 

\section{Intrinsic Evaluation}
In this section, we present intrinsic evaluation to assess the quantity and quality of extracted eventualities.

\subsection{Eventualities Extraction}

We first present the statistics of the extracted eventualities in Table~\ref{tab:eventuality-statistics}, which shows that simpler patterns like `s-v-o' appear more frequently than the complicated patterns like `s-v-be-a'.

\begin{figure}[t]
    \centering
    \includegraphics[width=0.9\linewidth]{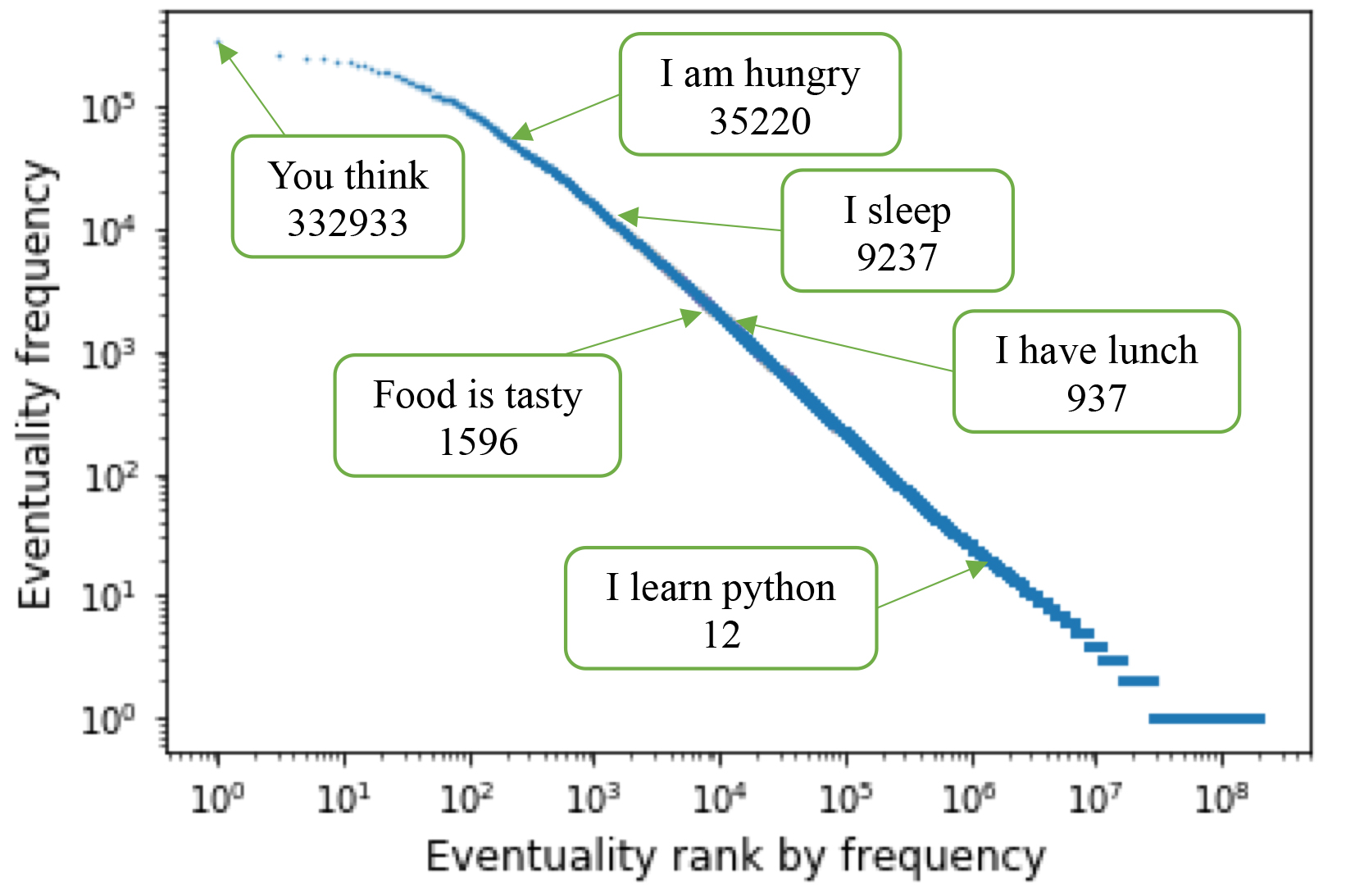}
    \caption{ Distribution of eventualities by their frequencies. }
    \label{fig:dist_eventualities}
\end{figure}

The distribution of extracted eventualities is shown in Figure~\ref{fig:dist_eventualities}.
In general, the distribution of eventualities follows the Zipf's law, where only a few eventualities appear many times while the majority of eventualities appear only a few times.
To better illustrate the distribution of eventualities, we also show several representative eventualities along with their frequencies and we have two observations.
First, eventualities which can be used in general cases, like `You think', appear much more times than other eventualities.
Second, eventualities contained in ASER are more related to our daily life like `Food is tasty' or `I sleep' rather than domain-specific ones such as `I learn python'.

After extracting the eventualities, we employ the Amazon Mechanical Turk platform (MTurk)\footnote{\url{https://www.mturk.com/}} for annotations. 
For each eventuality pattern, we randomly select 200 extracted eventualities and then provide these extracted eventualities along with their original sentences to the annotators.
In the annotation task, we ask them to label whether one auto-extracted eventuality phrase can fully and precisely represent the semantic meaning of the original sentence. 
If so, they should label them with `Valid'. Otherwise, they should label it with `Not Valid'.
For each eventuality, we invite four workers to label and if at least three of them give the same annotation result, we consider it to be one agreed annotation. Otherwise, this extraction is considered as disagreed.
In total, we spent \$201.6.
The detailed result is shown in Table~\ref{tab:eventuality-statistics}.
We got 2,597 agreed annotations out of 2,800 randomly selected eventualities, and the overall agreement rate is 92.8\%, which indicates that annotators can easily understand our task and provide consistent annotations.
Besides that, as the overall accuracy is 94.5\%, the result proves the effectiveness of the proposed eventuality extraction method.

\subsection{Relations Extraction}

\begin{figure*}[t]
    \centering
    \subfigure[Statistics and evaluation of bootstrapping.]{
    \label{fig:bootstrapping}
        \includegraphics[width=0.39\textwidth]{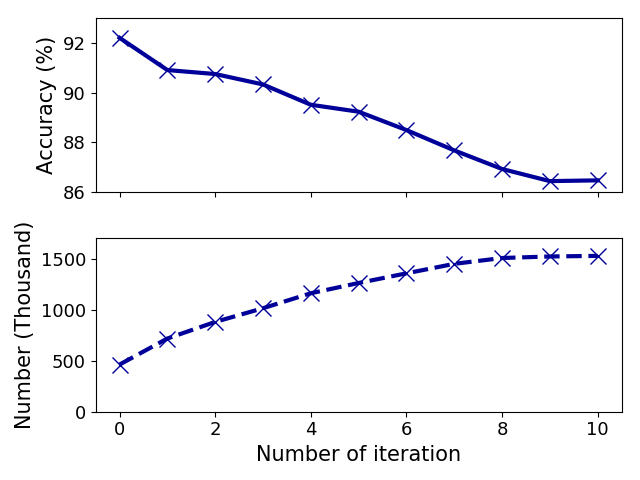}
    }
    \subfigure[Distribution and accuracy of different relation types.]{
    \label{fig:bootstrapping_result}
        \includegraphics[width=0.58\textwidth]{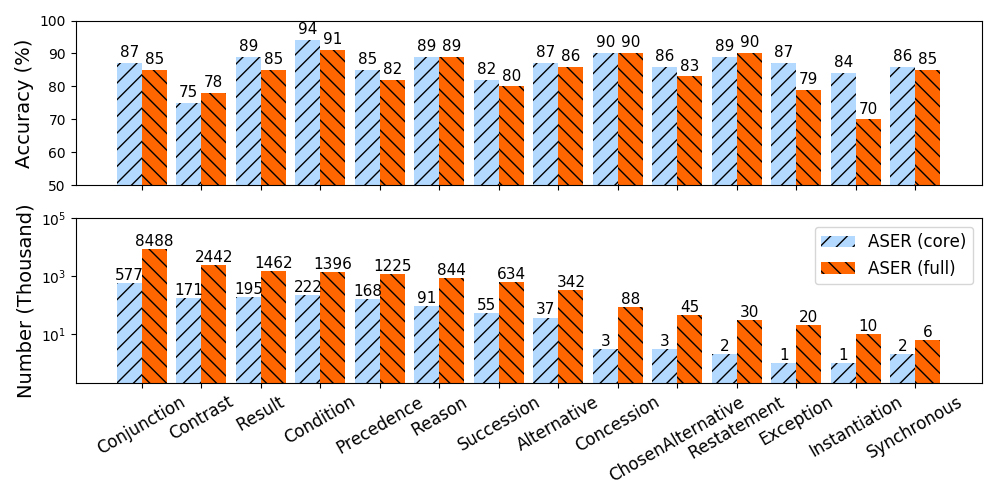}
    }
    \caption{Human Evaluation of the bootstrapping process. Relation \textit{Co\_Occurrence} is not included in the figures since it is not influenced by the bootstrapping. }\label{fig:relation_extraction}
\end{figure*}

In this section, we evaluate the quantity and quality of extracted relations in ASER. 
Here, to make sure the quality of the learned bootstrapping model, we filter out eventuality and eventuality pairs that appear once and use the resulting training instances to train the bootstrapping model.
The KG extracted from the selected data is called the core part of ASER. Besides that, after the bootstrapping, we directly apply the final bootstrapping model on all training instances and get the full ASER. In this section, we will first evaluate the bootstrapping process and then evaluate relations in two versions of ASER (core and full).

For the bootstrapping process, similar to the evaluation of eventuality extraction, we invite annotators from Amazon Turk to annotate the extracted edges. 
For each iteration, we randomly select 100 edges for each relation type.
For each edge, we generate a question by asking the annotators if they think certain relation exists between the two eventualities. If so, they should label it as `Valid'. Otherwise, they should label it as `Not Valid'.
Similarly, if at least three of the four annotators give the same annotation result, we consider it to be an agreed one and the overall agreement rate is 82.8 \%. 
For simplicity, we report the average accuracy, which is calculated based on the distribution of different relation types, as well as the total number of edges in Figure~\ref{fig:relation_extraction}(a).
The number of edges grows fast at the beginning and slows down later. After ten iterations of bootstrapping, the number of edges grows four times with the decrease of less than 6\% accuracy (from 92.3\% to 86.5\%). 

Finally, we evaluate the core and full versions of ASER.
For both versions of ASER, we randomly select 100 edges per relation type and invite annotators to annotate them using the same way as we annotating the bootstrapping process. 
Together with the evaluation on bootstrapping, we spent \$1698.4.
The accuracy along with the distribution of different relation types are shown in Figure~\ref{fig:relation_extraction}(b). 
We also compute the overall accuracy for the core and full versions of ASER by computing the weighted average of these accuracy scores based on the frequency. The overall accuracies of the core and full versions are 86.5\% and 84.3\% respectively, which is comparable with KnowlyWood~\cite{TandonMDW15KnowlyWood} (85\%), even though Knowlywood only relies on human designed patterns and ASER involves bootstrapping. 
From the result, we observe that, in general, the core version of ASER has a better accuracy than the full version, which fits our understanding that the quality of those rare eventualities might not be good.
But from another perspective, the full version of ASER can cover much more relations than the core version with acceptable accuracy.

\subsection{Comparison with ConceptNet}
\label{sec:auto-intrinsic-eval}
%In this section, we conduct intrinsic evaluation based on existing resources to demonstrate the quality of the ASER.

%\subsection{OMCS}

We study the relationship between ASER and the commonsense knowledge in ConceptNet~\cite{liu2004conceptnet}, or previously called Open Mind Common Sense (OMCS)~\cite{singh2002open}. The original ConceptNet contains 600K commonsense triplets and 75K among them involve eventualities, such as (\texttt{sleep}, \texttt{HasSubevent}, \texttt{dream}) and (\texttt{wind}, \texttt{CapableOf}, \texttt{blow to east}). 
All relations in ConceptNet are human-defined.
We select all four commonsense relations (\texttt{HasPrerequisite}, \texttt{Causes}, \texttt{MotivatedByGoal}, and \texttt{HasSubevent}) that involve eventualities to examine how many relations are covered in ASER.
%Here, we only consider those relations that has one eventuality in the head and one eventuality in the tail, such as <`sleep', HasSubevent, `dream'>.
Here by covered, we mean that for a given ConceptNet pair $(E_{o1}, E_{o2})$, we can find an edge $x=(E_{a1}, E_{a2}, c)$ in ASER such that $E_{o1}$ = $E_{a1}$, $E_{o1}$ = $E_{a1}$.
The detailed statistic of coverages are shown in Table~\ref{tab:OMCS-statistics}.

\begin{table}[t]
\centering
\caption{Statistics of selected OMCS data. we only select ConceptNet pairs that involve eventualities.}\label{tab:OMCS-statistics}	
{\footnotesize 
			\begin{tabular}{c|c|c|c}
				\toprule 
				Relation & \# Examples & \# Covered & Coverage \\
				\midrule
                HasPrerequisite     & 22,389 & 21,515 & 96.10\% \\
                Causes     & 14,065 & 12,605 & 89.62\% \\
                MotivatedByGoal     & 11,911 & 10,692 & 89.77\% \\
                HasSubevent     & 30,074 & 28,856 & 95.95\% \\
                \midrule
                Overall     & 78,439 & 73,668 & 93.92\% \\
				\bottomrule
			\end{tabular}
			}
\end{table}

\begin{figure}[h]
\centering
\includegraphics[width=\linewidth]{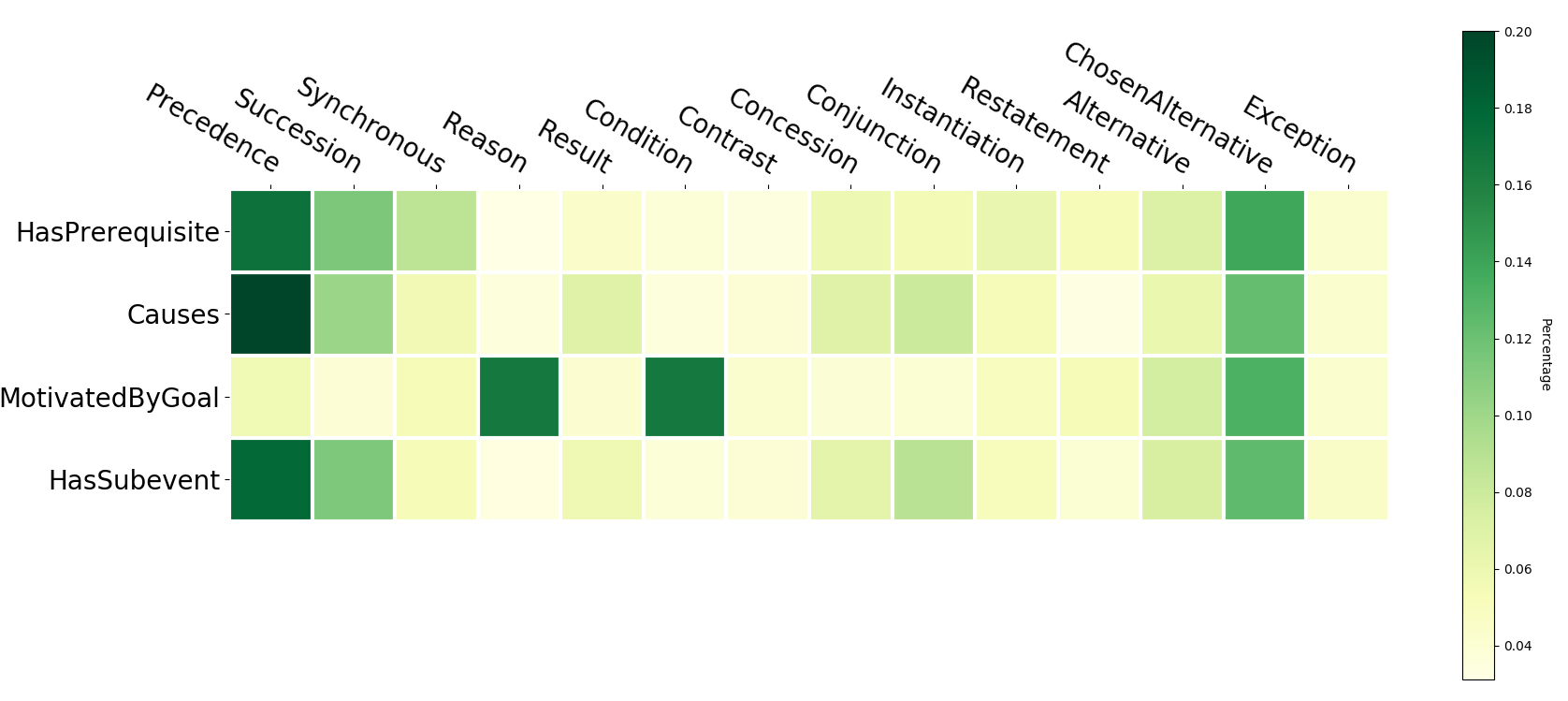}
\caption{\small Heatmap of overlapping between OMCS and ASER relations. For each OMCS relation, the distribution of matched ASER edges is computed. Darker color indicates more overlaps.}
\label{fig:heatmap}
\end{figure}

Moreover, to show the connection between the ConceptNet and ASER, we use a heat map to show the distribution of their relation pairs\footnote{As different relations are not evenly distributed, we normalize the co-occurrence number with the total number of ASER relations and then show it with the heatmap. } .
The result is shown in Figure~\ref{fig:heatmap}, where darker color indicates more coverage, and we observe many interesting findings. 
First, we can find a strong connection between \texttt{Causes} and \texttt{Precedence}, which fits our understanding that eventualities happen first is probably the reason for the eventualities happen later. 
For example, I eat and then I am full, where `we eat' is the reason of `I am full'.
Such correlation between temporal and causal relations is also observed in~\cite{ning2018joint}.
Second, most of the \texttt{MotivatedByGoal} pairs appear in the \texttt{Reason} or \texttt{Condition} relation in ASER, which makes sense because the motivation can be both the reason or the condition.
For example, `I am hungry' can be viewed as both the motivation and the requirement of `I eat' and both `I eat because I am hungry' and `I eat if I am hungry' are valid statements.
These observations demonstrate that the knowledge contained in ConceptNet can be effectively covered by ASER. Considering that ConceptNet is often criticized for its scale and ASER is two magnitude larger than ConceptNet, even though ASER may not be as accurate as ConceptNet, it could be a good supplement.

% most of the \texttt{MotivatedByGoal} pairs appear in the \texttt{Reason} relation in ASER, which fits our understanding that these two relations have the same meaning. 
% Notice that as ASER allows multiple relations between a pair of eventualities, we can also find a huge overlap between \texttt{Cause} relation \texttt{Precedence} relation, which also indicates the connection between temporal and causal relations.

% \todo[inline]{We'd better to have more discussion here to make it feels comprehensive: for each of the ConceptNet relation, we put some discussions. For some selected ASER relations, we put some discussions.}

\section{Extrinsic Evaluations}
In this section, we select the Winograd Schema Challenge (WSC) task to evaluate whether the knowledge in ASER can help understand human language.
Winograd Schema Challenge is known as related to commonsense knowledge and argued as a replacement of the Turing test~\cite{levesque2011Winograd}.
Given two sentences $s_1$ and $s_2$, both of them contain two candidate noun phrases $n_1$ and $n_2$, and one targeting pronoun $p$. The goal is to detect the correct noun phrase $p$ refers to. Here is an example~\cite{levesque2011Winograd}.

\noindent (1) The fish ate the worm. \texttt{It} was \texttt{hungry}.  Which was \texttt{hungry}?

 Answer: \texttt{the fish}.
 
\noindent (2) The fish ate the worm. \texttt{It} was \texttt{tasty}.  Which was \texttt{tasty}?

 Answer: \texttt{the worm}.

This task is challenging because $s_1$ and $s_2$ are quite similar to each other (only one-word difference), but the result is totally reversed. Besides that, all the widely used features such as gender/number are removed, and thus all the conventional rule-based resolution system failed on this task. For example, in the above example, both fish and worm can be hungry or tasty by themselves. 
In this experiment, we select all 273 questions\footnote{The latest winograd schema challenge contains 285 questions, but to be consistent with the baseline methods, we select the widely used 273 questions version.}.
% We can solve the problem because fish is subject of `eat' while the worm is the object, which requires understanding eventualities related to `eat'.
% Moreover, due to the small size of the Winograd schema challenge, supervised learning based methods are not practical.

% \subsection{Winograd Schema Challenge}
% brief introduction to Winograd

\subsection{ASER for Winograd Schema Challenge}

To demonstrate the effectiveness of ASER, we propose two methods of leveraging the knowledge in ASER to solve the WSC questions: (1) string match and inference over ASER; (2) Fine-tune pre-trained language models with knowledge in ASER.

\subsubsection{String Match and Inference}
For each question sentence $s$, we first extract eventualities with the same method introduced in Section~\ref{sec:eventuality-extraction} and then select eventualities $E_{n_1}$, $E_{n_2}$, and $E_p$ that contain candidates nouns $n_1$/$n_2$ and the target pronoun $p$ respectively. 
We then replace $n_1$, $n_2$, and $p$ with placeholder $X$, $Y$, and $P$,  and hence generate the pseudo-eventualities $E_{n_1}^\prime$, $E_{n_2}^\prime$, and $E_p^\prime$.
After that, if we can find the seed connectives in Table~\ref{tab:seed-pattern} between any two eventualities, we use the corresponding relation type as relation type $T$. Otherwise, we use \textit{Co\_Occurrence} as the relation type.  
To evaluate the candidate, we first replace the placeholder $P$ in $E_p^\prime$ with the corresponding placeholders $X$ or $Y$ and then use the following equation to define its overall plausibility score:
\begin{equation}
    F(n, p) = ASER_R(E_n^\prime, E_p^\prime),
\end{equation}
where $ASER_R(E_n, E_p)$ indicates the number of edges in ASER that can support that there exist one typed $T$ relation between the eventuality pairs $E_n^\prime$ and $E_p^\prime$.
For each edge ($E_h$, $T$, $E_t$) in ASER, if it can fit the following three requirements:
\begin{figure}[t]
    \centering
    \includegraphics[width=\linewidth]{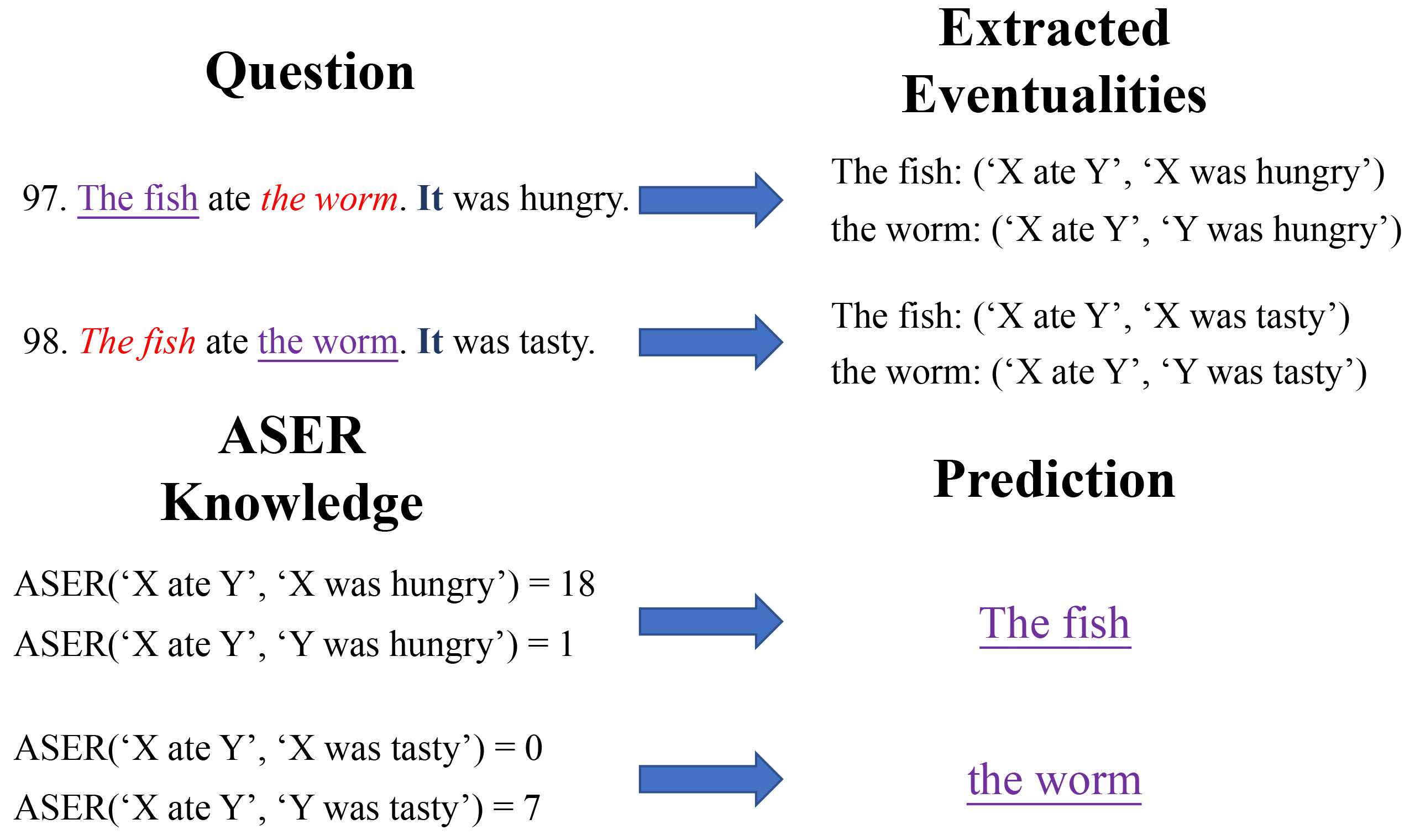}
    \caption{ Example of using ASER to solve Winograd questions. The number before questions are the original question ID. Correct answer and the other candidate are labeled with purple underline and red italic font respectively. }
    \label{fig:wino_case_study}
\end{figure}
\begin{enumerate}
    \item $E_h$ = $E_n^\prime$ other than the words in the place holder positions.
    \item $E_t$ = $E_p^\prime$ other than the words in the place holder positions.
    \item Assume the word in the placeholder positions of $E_h$ and $E_t$ are $w_h$ and $w_t$ respectively, $w_h$ has to be same as $w_t$.
\end{enumerate}
we consider that edge as a valid edge to support the observed eventuality pair.
If any of $E_n$ and $E_p$ cannot be extracted with our patterns, we will assign 0 to $F(n, p)$.
We then predict the candidate with the higher score to be the correct reference. If both of them have the same score (including 0), we will make no prediction.
At the current stage, We only use one-hop relations in ASER to perform the inference.
One example of using inference over ASER to solve WSC questions is shown in Figure~\ref{fig:wino_case_study}, our model can correctly resolve `it' to `fish' in question 97, because 18 edges in ASER support that the subject of `eat' should be `hungry', while only one edge supports the object of `eat' should be `hungry'. Similarly, our model can correctly resolve `it' to `the worm' in question 98, because seven edges in ASER support that the object of `eat' should be `tasty' while no edge supports that the subject of `eat' should be `tasty'.

\subsubsection{Fine-Tune Language Model}

Recently, Pre-trained Language models (e.g., BERT~\cite{DevlinCLT19} and GPT-2~\cite{radford2019language}) have demonstrated the strong ability to represent knowledge with deep models and have been proved to be very helpful for natural language understanding tasks like WSC. 
To evaluate if the knowledge in ASER can further help these language models, we propose to jointly leverage ASER and the language models to solve the WSC questions.

For each question sentence $s$, we first extract eventualities that contain the target pronoun $E_p$ and two candidates $E_c$ and then select all edges ($E_1$, $R$, $E_2$) from ASER such that $E_1$ and $E_2$ contains the verb of $E_p$ and $E_c$ respectively and there exists one common word in $E_1$ and $E_2$.
After that, for each selected edge, we then replace the common word in $E_2$ with a pronoun, which we assume refers to the same word in $E_1$, and thus create a pronoun coreference example in the WSC style.
For example, given the edge (`boy eats food', \textit{Reason}, `boy is hungry'), we create an example `boy eats food because he is hungry' and we know that `he' refers to `boy'.
Then, we fine-tune the pre-trained language models (BERT-large in our experiments) using these automatically generated data in the same way as~\cite{KCCYL2019}.
We denote this model as BERT+ASER.
In the original paper~\cite{KCCYL2019}, they also tried using another dataset WSCR~\cite{DBLP:conf/emnlp/RahmanN12} to fine-tune the language model, which is shown to be very helpful. Thus we also consider using both WSCR and ASER to fine-tune language models, which is denoted as BERT+ASER+WSCR. 

\subsection{Baseline Methods.} 
% To demonstrate the difficulty of the WSC, we first compare ASER with the state-of-the-art general co-reference resolutions:
% \begin{itemize}
% \item
% \textbf{Deterministic} model \cite{raghunathan2010multi}, which proposes one multi-pass seive model with human designed rules for the coreference resolution task.

% \item
% \textbf{Statistical} model~\cite{clark2015entity} uses human-designed entity-level features between clusters and mentions for coreference resolution.
% \item
% \textbf{Deep-RL} model~\cite{DBLP:conf/emnlp/ClarkM16} is a reinforcement learning method to directly optimize the coreference matrix instead of the traditional loss function.
% \item
% \textbf{End2end} model~\cite{DBLP:conf/naacl/LeeHZ18} is the current state-of-the-art coreference model, which performs in an end-to-end manner and leverages both the contextual information and a pre-trained language model~\cite{peters2018deep}.
% \end{itemize}

We first compare ASER with Conventional WSC solutions:
\begin{itemize}[leftmargin=*]
    \item \textbf{Knowledge Hunting}~\cite{DBLP:conf/emnlp/EmamiCTSC18} first search commonsense knowledge on search engines (e.g., Google) for the Winograd questions and then leverages rule-based methods to make the final predictions based on the collected knowledge.
    \item \textbf{LM} model~\cite{DBLP:journals/corr/abs-1806-02847} is the language model trained with very large-scale corpus and tuned specifically for the WSC task. We denote the best single language model and
the ensemble model as LM (single) and LM (ensemble) respectively.    % The performance of both single and ensemble language model are considered.
\end{itemize}

Besides that, we also compare with the selectional preference (SP) based method~\cite{zhang2019sp-10k}. Following the original setting, two resources (human annotation and Posterior  Probability) of SP knowledge are considered and we denote them as SP (human) and SP (PP) respectively\footnote{The original paper only considers two-hop SP knowledge for the WSC task, but we consider both one-hop and two-hop SP knowledge in our experiment.}.
Last but not least, we also compare with following pre-trained language models:
\begin{itemize}[leftmargin=*]
\item \textbf{GPT-2:} GPT-2~\cite{radford2019language} is the current best unsupervised model for the WSC task. We report the performance of its best model (1542 million parameters).
    \item \textbf{BERT:} As shown in~\cite{KCCYL2019}, we can first convert the original WSC task into a token prediction task and then BERT~\cite{DevlinCLT19} can be used to predict the results. As shown in the paper, the large model of BERT outperforms the base model significantly and we only report the performance of the large model. We denote the BERT model without or with fine-tuning on WSCR as BERT and BERT+WSCR respectively. 
    \end{itemize}
% \todo[inline]{Please also add results of SP-10K and PP based methods.}
The official released codes/models of all baseline methods are used to conduct the experiments.

\begin{table}[t]
    \centering
    \caption{ Experimental results on Winograd Schema Challenge. $\surd$ indicates the number of correct answers, $\times$ indicates the number of wrong answers, and $NA$ means that the model cannot give a prediction. $A_p$ means the prediction accuracy without $NA$ examples, and $A_o$ means the overall accuracy.}
    {\footnotesize
    \begin{tabular}{l|ccc|cc}
    \toprule
        
        Methods & $\surd$ & $\times$ & NA & $A_p$ & $A_o$ \\
    \midrule
        Random Guess & 137 & 136 & 0 & 50.2\% & 50.2\%  \\
        % \midrule
        % Deterministic~\cite{raghunathan2010multi}& 75 & 71 & 19 & 51.4\% & 51.2\% \\
        % Statistical~\cite{clark2015entity} & 75 & 78 & 12 & 49.0\% & 49.1\%  \\
        % Deep-RL~\cite{DBLP:conf/emnlp/ClarkM16} & 80 & 76 & 9 & 51.3\% & 51.2\% \\
        % End2end~\cite{DBLP:conf/naacl/LeeHZ18} & 79 & 84 & 2 & 48.5\% & 48.5\%  \\
        \midrule
        % USSM~\cite{liu2016combing} & 0 & 0 & 0 & 0\% & 52.8\% \\
        Knowledge Hunting~\cite{DBLP:conf/emnlp/EmamiCTSC18} & 119 & 79 & 75 & 60.1\% & 57.3\%  \\
        LM (single)~\cite{DBLP:journals/corr/abs-1806-02847} & 149 & 124 & 0 & 54.5\% & 54.5\%  \\
        LM (Ensemble)~\cite{DBLP:journals/corr/abs-1806-02847} & 168 & 105 & 0 & 61.5\% & 61.5\% \\
        \midrule
        SP (human)~\cite{zhang2019sp-10k} & 15 & 0 & 258 & \textbf{100\%} & 52.7\% \\
        SP (PP)~\cite{zhang2019sp-10k} & 50 & 26 & 197 & 65.8\% & 54.4\% \\
        \midrule
        GPT-2~\cite{radford2019language} & 193 & 80 & 0 & 70.7\% & 70.7\% \\
        BERT~\cite{KCCYL2019} & 169 & 104 & 0 & 61.9\% & 61.9\% \\
        BERT+WSCR~\cite{KCCYL2019} & 195 & 78 & 0 & 71.4\% & 71.4\% \\
        \midrule
        ASER (inference) & 63 & 27 & 183 & 70.0\% & 56.6\% \\
        BERT+ASER & 177 & 96 & 0 & 64.5\% & 64.5\% \\
        BERT+WSCR+ASER & 198 & 75 & 0 & 72.5\% & \textbf{72.5\%} \\
        % \midrule
        % Human Being & 0 & 0 & 0 & 0\% & 92\%\\
         \bottomrule
    \end{tabular}
    }
    \label{tab:wino}
\end{table}

\subsection{Result Analysis.}
From the result in Table~\ref{tab:wino}, we can make the following observations:
\begin{enumerate}[leftmargin=*]
    \item Pure knowledge-based methods (Knowledge Hunting and ASER (inference)) can be helpful, but their help is limited, which is mainly because of their low coverage and the lack of good application methods. For example, when we use ASER, we only consider the string match, which is obviously not good enough.
    \item Pre-trained language models achieve much better performance, which is mainly due to their deep models and large corpora they are trained on. Besides that, as already shown in~\cite{KCCYL2019}, fine-tuning on a similar dataset like WSCR can be very helpful.
    \item Adding knowledge from ASER can help state-of-the-art models. In our experiments, as we do not change any model architectures or hyper-parameters, we surprisingly find out that adding some related ASER knowledge can be helpful, which further proves the value of ASER.
\end{enumerate}

\section{Conclusions}\label{sec:conclusion}

In this paper, we introduce ASER, a large-scale eventuality knowledge graph.
We extract eventualities from texts based on the dependency graphs.
Then we build seed relations among eventualities using unambiguous connectives found from PDTB and
use a neural bootstrapping framework to extract more relations.
% \revisehm{Further bootstrapping the extraction to populate the relations to much more eventualities.
% We are the first approach using the large corpus to extract very large-scale eventuality KG. NOTE: both sentences are gramatically wrong and logically broken.}
ASER is the first large-scale eventuality KG using the above strategy.
We conduct systematic experiments to evaluate the quality and applications of the extracted knowledge.
Both human and extrinsic evaluations show that ASER is a promising large-scale eventuality knowledge graph with great potential in many downstream tasks.
% \revisehm{ASER and the related APIs will soon}

\section*{Acknowledgements}
This paper was supported by the Early Career Scheme (ECS, No. 26206717) from RGC in Hong Kong. We thank Dan Roth, Daniel Khashabi, and anonymous reviewers for their insightful comments on this work. We also thank Xinran Zhao for his help on the implementation of experiments on Winograd Schema Challenge.

\bibliography{main}

\bibliographystyle{ACM-Reference-Format}

\end{document}